\documentclass[review,12pt,3p]{elsarticle}

\usepackage{amssymb}
\usepackage{amsmath}
\usepackage{graphicx}
\usepackage{subcaption}
\usepackage{booktabs}
\usepackage{hyperref}

\usepackage[capitalize]{cleveref}
\crefname{section}{Sec.}{Secs.}
\Crefname{section}{Section}{Sections}
\Crefname{table}{Table}{Tables}
\crefname{table}{Tab.}{Tabs.}

\journal{Pattern Recognition}

\begin{document}

\begin{frontmatter}



\title{Heterophily-Aware Graph Attention Network}

\author[label1,label2,label3]{Junfu Wang}
\ead{wangjunfu@buaa.edu.cn}
\author[label1,label2]{Yuanfang Guo\corref{cor1}}
\ead{andyguo@buaa.edu.cn}
\author[label4,label5]{Liang Yang}
\ead{yangliang@vip.qq.com}
\author[label2]{Yunhong Wang}
\ead{yhwang@buaa.edu.cn}
\affiliation[label1]{organization={State Key Laboratory of Software Development Environment, Beihang University},
            city={Beijing},
            postcode={100191},
            country={China}}

\affiliation[label2]{organization={School of Computer Science and Engineering, Beihang University},
            city={Beijing},
            postcode={100191},
            country={China}}

\affiliation[label3]{organization={Shen Yuan Honors College, Beihang University},
            city={Beijing},
            postcode={100191},
            country={China}}

\affiliation[label4]{organization={School of Artificial Intelligence, Hebei University of Technology},
            city={Tianjin},
            postcode={300401},
            country={China}}

\affiliation[label5]{organization={Hebei Province Key Laboratory of Big Data Calculation, Hebei University of Technology},
            city={Tianjin},
            postcode={300401},
            country={China}}

\cortext[cor1]{Corresponding author.}



\begin{abstract}
    Graph Neural Networks (GNNs) have shown remarkable success in graph representation learning.
    Unfortunately, current weight assignment schemes in standard GNNs, such as the calculation based on node degrees or pair-wise representations, can hardly be effective in processing the networks with heterophily, in which the connected nodes usually possess different labels or features.
    Existing heterophilic GNNs tend to ignore the modeling of heterophily of each edge, which is also a vital part in tackling the heterophily problem.
    In this paper, we first propose a heterophily-aware attention scheme and reveal the benefits of modeling the edge heterophily, i.e., if a GNN assigns different weights to edges according to different heterophilic types, it can learn effective local attention patterns, enabling nodes to acquire appropriate information from distinct neighbors.
    Then, we propose a novel Heterophily-Aware Graph Attention Network (HA-GAT) by fully exploring and utilizing the local distribution as the underlying heterophily, to handle the networks with different homophily ratios.
    To demonstrate the effectiveness of the proposed HA-GAT, we analyze the proposed heterophily-aware attention scheme and local distribution exploration, by seeking an interpretation from their mechanism.
    Extensive results demonstrate that our HA-GAT achieves state-of-the-art performances on eight datasets with different homophily ratios in both the supervised and semi-supervised node classification tasks.
\end{abstract}




\end{frontmatter}


\section{Introduction}
\label{Sec-intro}

Due to their success in modeling the irregular graph data, Graph Neural Networks (GNNs) \cite{gcn,gat,graphsage,bi-gcn,bai2021learning,cui2021learning,regnn,bigcn2} have been applied to various tasks, including social analysis \cite{gnn_app_social1}, computer vision \cite{gnn_app_object_tracking1}, and natural language processing \cite{nlp2}.
Current GNNs mostly follow the Message Passing strategy \cite{message-passing}, where the node state is iteratively updated by interacting with its neighboring nodes.
Existing approaches mainly differ in the node aggregation mechanisms, which describe how each node aggregates the representations of its neighbors with its own ones by adopting different weight assignment schemes.

Therefore, one of the critical interests in GNN research is the design of a powerful weight assignment scheme, which can effectively determine the importance of different representations in every node and its neighbors.
Graph Convolutional Network (GCN) \cite{gcn} and its variants \cite{graphsage,gtn,wijesinghe2021new} utilize the structural information, e.g., symmetric normalization with respect to the node degree, as the edge weight. Besides, Graph Attention Network (GAT) \cite{gat} generates an attention score (weight) by utilizing the self-attention mechanism.
It allows every node to compute an attention score of its neighbors and softly select its most relevant neighbors.
Several attention-based weight assignment schemes are subsequently proposed by employing different pair-wise attention functions \cite{GAT-cosine,Gaan,SuperGAT,CS-GNN,gatv2}.
Typically, these two kinds of weight assignment schemes can achieve state-of-the-art results on the networks, which satisfy the homophilic assumption, i.e., nodes prefer to connect with other nodes possessing the same label.
Unfortunately, recent studies \cite{H2GCN,geom-gcn,DMP} have shown that the above mentioned weight assignment schemes cannot perform well when the homophilic assumption is not satisfied, e.g., on heterophilic networks, in which the linked nodes usually possess different labels or attributes.

Several improvements have been proposed to tackle the heterophily problem by designing adaptive weight assignment schemes.
From the spectral perspective, FAGCN \cite{FAGCN} adopts a self-gating attention mechanism to assign weights to the coefficients for both the low-frequency and high-frequency signals.
DMP \cite{DMP} considers attribute heterophily and specifies every attribute propagation weight on each edge.
Unfortunately, these methods have ignored to model the edge heterophily, which is determined by the type of the connected nodes for each edge, according to the labels or attributes.
This is also critical for solving the heterophily problem.
BM-GCN \cite{BM-GCN} attempts to estimate the label heterophily directly and assigns weights based on the block similarity. 
Unfortunately, it assigns higher weights to the edges connecting similar nodes, which is insufficient to exploit information from the dissimilar nodes.

In this paper, we first propose an intuitive solution by constructing a GNN, which can assign different weights to edges according to different heterophilic types.
Then, the GNN can learn an effective local attention pattern, which enables nodes to acquire appropriate information from both similar and dissimilar nodes.
For verification, we develop a heterophily-aware attention scheme to handle the graph with a prior edge heterophily for every edge, which is specified by the labels of its connected nodes.
By modeling the importance of different edge heterophilic types via certain adaptive weights, we empirically validated that this GNN can excel in the node classification tasks, even on the heterophilic graphs.
For example, on the Actor and Chameleon datasets, it can achieve 63.38/33.31 and 92.27/67.96 accuracies with/without the adaptive weights for different heterophilic types, respectively.
This observation indicates that the edge heterophily plays a crucial role in exploiting information from both the similar and dissimilar nodes, and thus solves the heterophily problem.

Unfortunately, in real-world node classification tasks, the edge heterophily cannot be directly computed when the majority of the node labels are not given, especially in the semi-supervised settings.
Thus, the remaining challenge is how to acquire proper heterophily information.
Previous studies \cite{BM-GCN,cpgnn} attempt to estimate the edge heterophily according to the node labels by a pre-trained MLP.
Unfortunately, the precise node labels are hard to be directly predicted with the pre-trained MLP.

Recent studies \cite{ma2021homophily,wang2024understanding,MWGNN} have shown that the local distribution, which describes the distribution measured by certain characteristics in the local neighborhood, plays a critical role in handling the heterophilic graphs. 
According to \cite{ma2021homophily,wang2024understanding}, GCN can outperform MLP on the heterophilic graphs, in which the local distributions computed by the node labels are distinctive, i.e., nodes in the same classes tend to possess similar local distributions while nodes from different classes tend to possess dissimilar ones.
Meanwhile, according to \cite{MWGNN}, the topological structure, node features, and positional identity are also decisive for estimating effective local distributions and then facilitating the weight assignment scheme.

In this paper, we tackle the heterophily problem by proposing a novel Heterophily-Aware Graph Attention Network (HA-GAT), which can handle graphs with different homophily ratios.
Instead of directly estimating the heterophily from the node labels, we construct an explorer network to excavate local distributions to represent the underlying heterophily of nodes.
The heterophily preference of each edge is derived by integrating the underlying categories of the connected nodes.
Then, according to our heterophily-aware attention scheme, the heterophily preference is utilized to generate the attention coefficients.
Specifically, a learnable parsing matrix with a gradient scaling factor is constructed in each layer, to compute with the heterophily preference matrices and generate the attention coefficients.
Each element in the parsing matrix represents the corresponding importance for a particular underlying heterophilic edge type, which can be learned adaptively. 
At last, the normalized attention coefficients are utilized in the message aggregation process.

To demonstrate the effectiveness of the proposed HA-GAT, we analyze the proposed heterophily-aware attention scheme and local distribution exploration, by seeking an interpretation from their mechanism.
Extensive experiments on eight datasets with different homophily ratios demonstrate the effectiveness of our proposed HA-GAT, for both the supervised and semi-supervised node classification tasks.
Our major contributions are summarized as follows:
\begin{itemize}
  \item We empirically reveal the benefits of modeling the edge heterophily, i.e., if a GNN assigns different weights to edges according to different heterophilic types, it can learn effective local attention patterns, enabling nodes to acquire appropriate information from distinct neighbors.
  \item We propose a heterophily-aware attention scheme to adaptively assign weights to different heterophilic types, by fully exploiting the given edge heterophily.
  \item We propose a novel Heterophily-Aware Graph Attention Network (HA-GAT) by fully exploring and utilizing the local distribution as the underlying heterophily, to handle the networks with different homophily ratios.
\end{itemize}

\section{Related Work}

\subsection{Attention-based Graph Neural Networks}
Graph Attention Network (GAT) \cite{gat} pioneers to compute the hidden representation of each node by attending over its neighbors, based on a self-attention strategy. Specifically, it generates attention coefficients by performing an inner product between a learnable vector and the concatenated representations of two connected nodes. Subsequently, several GAT variants are proposed by different calculation schemes of the attention coefficients, such as cosine similarity \cite{GAT-cosine} and element-wise product \cite{Gaan,SuperGAT}.
Besides, some methods are proposed to improve GAT from other perspectives.
CS-GNN \cite{CS-GNN} improves GAT by predicting the attention coefficients based on separate scoring representations, instead of using direct node representations.
SuperGAT \cite{SuperGAT} proposes a self-supervised strategy to learn the attention coefficients with respect to the edge information.
GATv2 \cite{gatv2} argues that the attention scores learned by GAT \cite{gat} are irrelevant to the ego-representation. Then, it presents an improved variant by modifying the order of operations.
All of these attention-based GNNs function decently under the homophilic assumption, i.e., they attempt to measure similarities or distances by employing distinct pair-wise attention functions, which makes them unsuitable to process the networks with heterophily.

\subsection{GNNs for Heterophily}
Recently, several methods have been proposed to tackle the heterophily problem, which can be divided into two lines.

The first line of approaches intends to model the diverse local patterns via designing adaptive message aggregation schemes.
Geom-GCN \cite{geom-gcn} builds a structural neighborhood by leveraging network embeddings.
The neighborhood is then partitioned into several geometric blocks and the neighbors within each block are aggregated separately.
FAGCN \cite{FAGCN} adopts a self-gating attention mechanism to learn the attention coefficients for both the low-frequency and high-frequency signals.
Besides, ACM-GNN \cite{acm-gnn} proposes an Adaptive Channel Mixing framework to adaptively exploit low-frequency, high-frequency, and identity channels.
GBK-GNN \cite{gbkgnn} employs a bi-kernel feature transformation and a selection gate to simultaneously model the similarity and dissimilarity (i.e., the positive and negative correlations) between node features.
By taking node attributes as weak labels, DMP \cite{DMP} considers the attribute heterophily for diverse message passing and specifies every attribute propagation weight on each edge.
WRGNN \cite{WRGNN} transforms a heterophilic input graph into a multi-relational graph, in which the proximity and structural information are represented as distinct types of edges.
BM-GCN \cite{BM-GCN} leverages a two-staged block-guided classified aggregation mechanism, to directly model the explicit heterophily and assign edge weights according to the block similarities.
MWGNN \cite{MWGNN} utilizes well-designed  handcrafted features, i.e., local degree profile, sorted node feature, and shortest path distance, to predict the underlying local distribution, and assigns the edge weights by a topology-attribute decoupling mechanism.
Although it also considers the local distribution, handcrafted features may not be appropriate for handling all the graphs.
Different from the existing literature, our proposed HA-GAT estimates the underlying heterophilic type of each edge to avoid utilizing imprecise label information. It leverages a heterophily-aware attention scheme to adaptively learn the importance of each heterophilic type, presenting a novel paradigm for addressing heterophily.

The other line of approaches focuses on expanding the local neighborhood.
H2GCN \cite{H2GCN} explicitly aggregates information from high-order neighborhoods to find more similar neighbors, which has proved to be beneficial in heterophily settings.
This design is also employed in \cite{gcn2,mixhop}.
GRP-GNN \cite{gpr-gnn} proposes a generalized PageRank architecture to model the importance of different orders of neighborhoods adaptively.
Besides, some works utilize different ranking strategies to find potential neighbors.
Non-local GNN \cite{non-local-gnn} designs an attention-guided sorting strategy to find distant but informative nodes and put them together.
GPNN \cite{GPNN} leverages a pointer network to select the most relevant nodes from a large amount of multi-hop neighborhoods.
UGCN \cite{UGCN} introduces a discriminative aggregation to fuse 1-hop, 2-hop, and $k$NN neighbors simultaneously.
GloGNN \cite{glognn} generates node embeddings by aggregating information from global nodes within the graph.
Our HA-GAT follows the first line and thus is not similar to these methods.

\section{Preliminaries}

\subsection{Notations}
An attributed graph is denoted as $\mathcal{G}=\left\{\mathcal{V},\mathcal{E},\mathcal{X}\right\}$ with the vertex set $\mathcal{V} =\{v_i\}_{i=1}^N$ and edge set $|\mathcal{E}|= E$. 
Each node $v_i$ contains a feature $x_i\in\mathbb{R}^d$.
$X\in\mathbb{R}^{N\times d}$ is the collection of all the features in all the nodes. 
$A=[a_{ij}] \in\left\{0, 1\right\}^{N\times N}$ is the adjacency matrix, where $a_{ij}=1$ if nodes $v_i$ and $v_j$ are connected, or $a_{ij}=0$ otherwise.
$d_i=\sum_j a_{ij}$ stands for the degree of node $v_i$ and
$D=diag(d_1, d_2, \dots, d_n)$ represents the degree matrix corresponding to the adjacency matrix $A$.
For each node $v_i$, $\mathcal{N}_i = \left\{v_j | a_{ij}=1 \right\}$ represents the set of its neighbors and $\tilde{\mathcal{N}}_i = \mathcal{N}_i\cup\left\{v_i\right\}$ denotes the set containing both the node $v_i$ and its neighbors.

\subsection{Homophily versus Heterophily}
To clearly measure the homophily/heterophily of a graph, we follow \cite{geom-gcn} to define the homophily ratio $\mathcal{H}(\mathcal{G})$ as
\begin{equation}
  \mathcal{H}(\mathcal{G}) = \frac{1}{V}\sum_{v_i\in \mathcal{V}} \frac{\sum_{v_j\in\mathcal{N}_i}(y_i = y_j)}{|\mathcal{N}_i|},
  \label{eq-homophily_ratio}
\end{equation}
where $\mathcal{H}(\mathcal{G})\in \left[0,1\right]$.
A high homophily ratio indicates that the graph is with strong homophily, while a graph with strong heterophily possesses a small homophily ratio.

\section{Heterophily-Aware Attention Scheme}
\label{sec-attention}

This section develops a heterophily-aware attention scheme to adaptively assign weights by fully exploiting the given edge heterophily.

We first assume that the edge heterophily preference matrix $\mathbf{m_{ij}}$ for each edge is acquirable, with respect to the node labels. It can be computed via
\begin{equation}
  \mathbf{m_{ij}} = y_i \otimes y_j,
\end{equation}
where $\mathbf{m_{ij}}\in\{0,1\}^{C\times C}$ represents the distribution among $(C\times C)$ heterophilic types for edge $e_{ij}$, and $\otimes$ represents the outer product of two vectors.
Note that $y_i\in\mathbb{R}^C$ is a one-hot vector, which represents the ground-truth label of node $v_i$.
The element in $\mathbf{m_{ij}}$, $m_{ijk_ik_j} = y_{ik_i}\cdot y_{jk_j}$, is 1 if and only if the nodes $v_i$ and $v_j$ belong to classes $k_i$ and $k_j$, respectively, and 0 otherwise.

Then, the heterophily preference of each edge, $\mathbf{m_{ij}}$, is interpreted by a learnable parsing matrix to generate the attention coefficient in each layer.
In the $l$-th layer, the parsing matrix $\mathbf{\omega}^{(l)}\in\mathbb{R}^{C\times C}$ is employed as a parameter matrix, which can be optimized in the training process.
Then, the heterophily-aware attention coefficient for each edge $e_{ij}$ can be predicted via
\begin{equation}
  w_{ij}^{(l)} = \langle \mathbf{m_{ij}}, \phi(\mathbf{\omega}^{(l)}) \rangle_F = tr\left(\mathbf{m_{ij}}^T\phi(\mathbf{\omega}^{(l)})\right),
  \label{eq-w-1}
\end{equation}
where $\langle\cdot ,\cdot \rangle_F$ denotes the Frobenius inner product, which is an inner product operation in the matrix domain, and $tr(\cdot)$ is the trace of the matrix. 
Here, $\phi(\cdot)$ is an element-wise computing function to calculate the importance of each heterophilic edge type with respect to the real-valued parameters.
For each parameter $\omega^{(l)}$ in $\mathbf{\omega}^{(l)}$, 
\begin{equation}
  \phi(\omega^{(l)}) = \left(\lambda \omega^{(l)}\right)_{+} = \max \left\{ \lambda \omega^{(l)}, 0 \right\},
  \label{eq-phi}
\end{equation}
where $\lambda > 0$ is a gradient scaling factor, which is employed to enlarge/shrink the gradient of each $\omega^{(l)}$.
Specifically, each parameter $\omega^{(l)}$ is initially set as $\frac{1}{\lambda}$ to ensure $\phi(\lambda\cdot\omega^{(l)})$ being one, i.e., each heterophilic type of edges possesses equivalent importance at the beginning of the training process.
With the gradient scaling factor $\lambda$, the modification of $\lambda\mathbf{\omega}^{(l)}$ in the optimization step will be scaled.
Besides, the $\left(\cdot\right)_{+}$ function is utilized to ensure the weights to be non-negative.
During the training process, $\phi(\omega^{(l)})$ gradually becomes diverse, each of which reflects a corresponding importance for a certain heterophilic edge type.

Typically, the self-loop connection, which is a special type of edge, is always taken into consideration for combining the representations of the central node and its neighbors.
Thus, a parameter $\omega_{sl}^{(l)} \in \mathbb{R}$ is utilized here to model the importance of the self-loop connection, as
\begin{equation}
  w_{ii}^{(l)} = \phi \left(\omega_{sl}^{(l)} \right).
\end{equation}

Note that $\left\{\phi(\mathbf{\omega}^{(l)}), \phi (\omega_{sl}^{(l)})\right\}$ determines how each type of nodes assign weights to their neighbors and themselves.
We denote it as a Local Attention Pattern (LAP) to facilitate the subsequent analysis.

\noindent\textbf{Neighbor Norm.}
Different from the commonly utilized normalization methods, we construct a Neighbor Norm to normalize the attention coefficients, as
\begin{equation}
  \alpha_{ij}^{(l)} = \frac{w_{ij}^{(l)}}{\sum_{v_k\in\tilde{\mathcal{N}}_j} w_{jk}^{(l)}},
  \label{eq-hagat-norm}
\end{equation}
where $\tilde{\mathcal{N}}_j = \mathcal{N}_j \cup\left\{v_j\right\}$ is the set containing node $v_j$ and its neighbors.
Each attention coefficient $w_{ij}^{(l)}$ of edge $e_{ij}$ is normalized by the weighted degree of the neighbor node $v_j$.
This design assumes that the connections to general neighbors (i.e., neighbors with small weighted degrees) are more responsive to the characteristics of the node than the connections to popular neighbors (i.e., neighbors with large weighted degrees).
This assumption is consistent with various real-world networks, e.g., social networks and citation networks.

After obtaining the normalized attention coefficients, the message aggregation step can be performed as
\begin{equation}
  h^{(l+1)}_i = \sigma\left(\sum_{v_j\in\tilde{\mathcal{N}}_i} \alpha_{ij}^{(l)} h_j^{(l)}\Theta^{(l)}\right),
  \label{eq-aggregation}
\end{equation}
where $\Theta^{(l)}$ is a parameter matrix and $\sigma$ is the ReLU function.

\noindent\textbf{Remark.}
Our heterophily-aware attention scheme utilizes one parameter to model the importance of each heterophilic edge type.
It attempts to find an effective LAP to help nodes extract appropriate information from both the similar and dissimilar neighbors.

\begin{figure*}[t]
  \centering
  \includegraphics[width=\columnwidth]{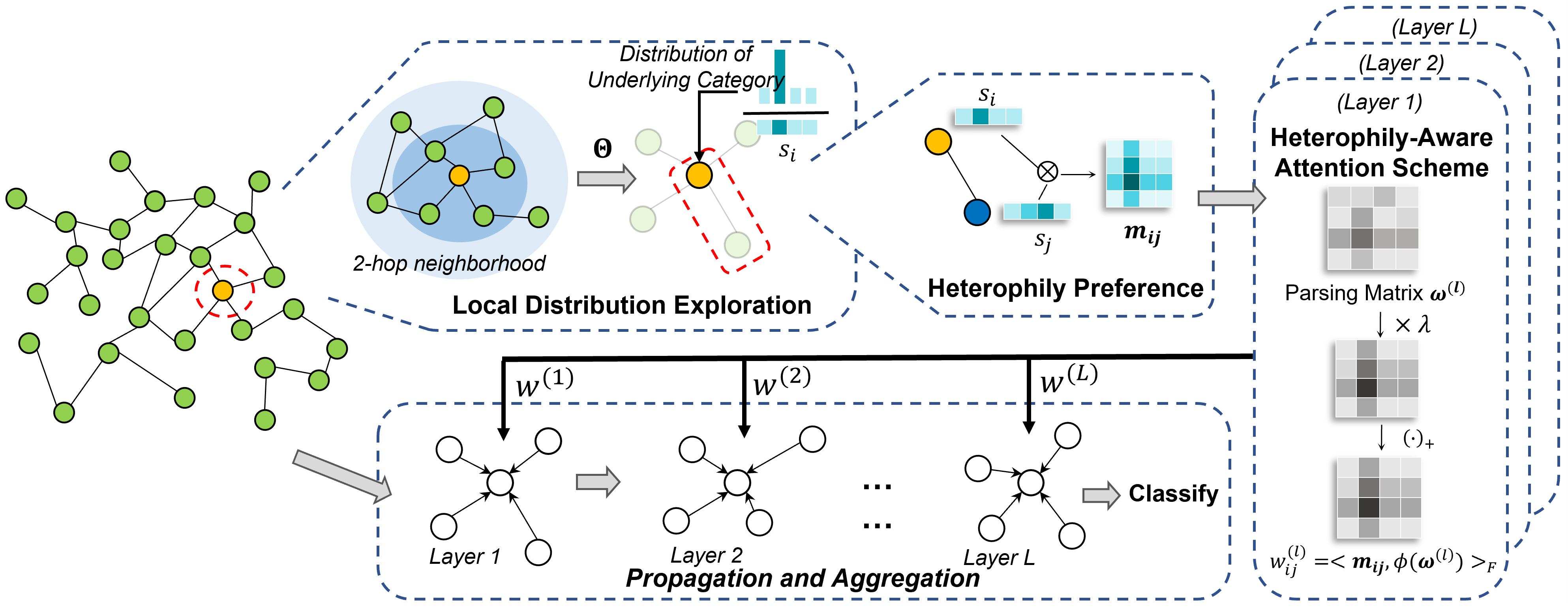}
  \caption{Illustration of our Heterophily-Aware Graph Attention Network (HA-GAT).}
  \label{fig-framework}
\end{figure*}

\section{Heterophily-Aware Graph Attention Network}
The heterophily-aware attention scheme proposed in \cref{sec-attention} can adaptively assign weights based on the given edge heterophily, thereby addressing the heterophily issue. Unfortunately, in real-world node classification tasks, the edge heterophily with respect to node labels cannot be directly computed when the majority of node labels are unavailable, especially in semi-supervised settings.

To address this challenge, this section introduces a Heterophily-Aware Graph Attention Network (HA-GAT) that fully explores and utilizes the local distribution as underlying heterophily. Subsequently, we compare our HA-GAT with typical attention-based GNNs and elucidate its connections to existing GNNs, for a deeper understanding. A comprehensive illustration of our HA-GAT is provided in \cref{fig-framework}.

\subsection{Local Distribution Exploration}
Local distribution describes a distribution measured by certain characteristics in each local neighborhood.
It can be measured by the node label \cite{ma2021homophily} as well as other local characteristics \cite{MWGNN}.
In general, the local distribution of each node $v_i$ can be derived from its local neighborhood, via
\begin{equation}
  s_i = \Theta\left(x_i, \{x_j\}_{j\in \mathcal{N}_i^{(p)}}\right),
  \label{eq:s_i}
\end{equation}
where $s_i \in \mathbb{R}^{t}$ represents the local distribution of node $v_i$ and $\mathcal{N}_i^{(p)}$ stands for the collection of nodes within a $p$-hop local neighborhood.
Note that $t>0$ is a preset integer constant which represents the number of underlying categories.
Each element in $s_i$ ($s_{ik}\geq 0$) represents the probability of assigning category $k$ to node $v_i$, and $\sum_k s_{ik} = 1$.
Note that $\Theta(\cdot)$ is expected to estimate the appropriate local distribution, according to certain underlying measurements, and thus facilitates the node classification task.

In this paper, instead of generating local distribution by directly estimating the node labels or from well-designed handcrafted features \cite{MWGNN}, we excavate local information from the local smoothed features.
This design is motivated by two aspects.
Firstly, based on the analysis \cite{ma2021homophily} under certain ideal assumptions, the local smoothed features tend to contribute similarly in representing heterophily, compared to the local distributions.
Under such circumstances, GCN can perform decently by obtaining local smoothed features, because the local distributions according to the node labels are discriminative.
Secondly, in practical situations, \cite{MWGNN} empirically verifies that the local distributions, which are generated from the local node features and the topological information, can facilitate the weight assignment scheme.
However, instead of generating the local distributions based on the handcrafted characteristics from the neighborhoods, e.g., local degree profile, the local smoothed features can be utilized to adaptively select appropriate characteristics as underlying measurements.

Therefore, a 2-layered GCN is employed as an explorer network to excavate local distributions from the local smoothed features as the underlying heterophily.
Then, \cref{eq:s_i} can be rewritten as 
\begin{equation}
  S = {\rm softmax}\left( \tilde{A} \ \sigma \left( \tilde{A} X W^{(0)}_e\right) W^{(1)}_e\right),
  \label{eq:s_i-GCN}
\end{equation}
where $S = \{s_1; s_2; ...; s_N\} \in \mathbb{R}^{N\times t}$ represents the local distributions.
As stated above, we assume that the local distribution reflects the corresponding node categories, i.e., it can be transformed to the category distribution by a certain linear transformation $W_{t}$.
Thus, we can utilize $S$ to approximately model the underlying heterophily by combining the linear projections $W^{(1)}_e$ and $W_{t}$.
Then, $s_i$ represents the category distribution among the underlying categories of node $v_i$.
Note that $\sigma(\cdot)$ is the ReLU function, $\tilde{A}=\hat{D}^{-\frac{1}{2}}\hat{A}\hat{D}^{-\frac{1}{2}}$, $\hat{A}=A+I$, and $\hat{D}$ is the degree matrix of $\hat{A}$.

Then, the underlying heterophily preference of each edge is re-computed via
\begin{equation}
  \mathbf{m_{ij}} = s_i\otimes s_j,
\end{equation}
where $\mathbf{m_{ij}}\in\mathbb{R}^{t\times t}$ represents the probability distribution of the $(t\times t)$ heterophilic types for edge $e_{ij}$.
Each element $m_{ijk_ik_j}$ in $\mathbf{m_{ij}}$, where $m_{ijk_ik_j} = s_{ik_i}\cdot s_{jk_j}$, represents the heterophily preference that edge $e_{ij}$ connects two nodes with their categories being $k_i$ and $k_j$, respectively.
Since $\sum_k^{t}s_{ik}=1$ and $\sum_k^{t}s_{jk}=1$, $\mathbf{m_{ij}}$ is also a probability matrix, i.e., $\sum_{k_1}^t\sum_{k_2}^t m_{ijk_1k_2}= 1$.
Therefore, $\mathbf{m_{ij}}$ indicates the heterophily preference of each edge $e_{ij}$, which is utilized in the heterophily-aware attention scheme to generate the attention coefficients and constructs our HA-GAT.
Note that we denote a special variant of our HA-GAT with a prior edge heterophily according to the node labels as HA-GAT(L), which is introduced in \cref{sec-attention}.

\subsection{Discussions with Existing GNNs}
Here, we compare the proposed HA-GAT with typical attention-based GNNs, and analyze its connections to existing GNNs by proposing several HA-GAT variants.

\subsubsection{Comparisons with Existing Attention-based GNNs}
Currently, the attention-based GNNs \cite{gat,gatv2,CS-GNN,SuperGAT} usually exploit a self-attention strategy, where the attention coefficients on the edges are correlated to the representations of their correspondingly connected nodes, i.e.,
\begin{equation}
  w_{ij}^{(l)} =\Phi_s^{(l)}\left(h_i^{(l-1)}, h_j^{(l-1)}\right),
  \label{eq-self-attention-strategy}
\end{equation}
where $w_{ij}^{(l)}$ is the attention coefficient of edge $e_{ij}$ in the $l$-th layer and $h_i^{(l)}$ is the representations of node $v_i$ in the $l$-th layer.
Note that $\Phi_s^{(l)}(\cdot)$ is a well-designed attention function, which is designed differently in the existing methods \cite{gat,gatv2,GAT-cosine}.
Then, the attention coefficient is normalized by a ${\rm softmax}$ function \cite{gat,gatv2}, i.e., 
\begin{equation}
  \alpha_{ij}^{(l)} = {\rm softmax}\left(w_{ij}\right) = \frac{{\rm exp}(w_{ij})}{\sum_{v_k\in\tilde{\mathcal{N}_i}}{\rm exp}(w_{ik})}.
  \label{eq-gat-norm}
\end{equation}

On the contrary, our HA-GAT calculates the underlying heterophily preference $\mathbf{m_{ij}}$ for each edge, which is correlated to the local neighborhood in the attributed graph, i.e.,
\begin{equation}
  \mathbf{m_{ij}} =\Phi_u\left(x_i, x_j, \{x_{k_i}\}_{k_i\in \mathcal{N}_i^{(2)}}, \{x_{k_j}\}_{k_j\in \mathcal{N}_j^{(2)}}\right),
  \label{eq-local-strategy-1}
\end{equation}
where $\Phi_u(\cdot)$ calculates the heterophily preference $\mathbf{m_{ij}}$ from the local neighborhood.
Then, $\mathbf{m_{ij}}$ is utilized to compute the attention coefficients in each layer as
\begin{equation}
  w_{ij}^{(l)} = f^{(l)}(\mathbf{m_{ij}}),
  \label{eq-local-strategy-2}
\end{equation}
where $f^{(l)}(\cdot)$ is the parsing function in the $l$-th layer to be utilized to generate the heterophily-aware attention coefficient from the heterophily preference matrix.

Compared to \cref{eq-self-attention-strategy,eq-local-strategy-1}, our $\Phi_u(\cdot)$ fully excavates information in the local neighborhood, while $\Phi_s^{(l)}(\cdot)$ only measures the attention coefficient based on the pair-wise node representations.
Although these pair-wise node representations contain information of local neighborhood when $l\geq 1$, they are task-specific, i.e., they are restricted to maintain the classification information.
Thus, their correspondingly calculated attention coefficients can hardly be optimal without fully considering the local neighborhood.

To clearly demonstrate the effectiveness of our attention scheme, we develop 
1) a GAT-like HA-GAT variant, i.e., HA-GAT(G), by utilizing the representation in each HA-GAT layer to generate $\mathbf{m_{ij}}$,
and 2) a MLP based HA-GAT variant, i.e., HA-GAT(M), by replacing the 2-layered GCN with a 2-layered MLP as the explorer network.

\subsubsection{Variants and Connections to existing GNNs}
To better understand our HA-GAT, we discuss its connections to the existing GNNs by constructing three HA-GAT variants.

(1) When the underlying category dimension $t$ is set to 1, the calculation of the attention coefficient in \cref{eq-w-1} will degenerate to
\begin{equation}
  w_{ij}^{(l)} = \phi\left(\omega\right),
\end{equation}
where $\omega\in\mathbb{R}$ is a parameter value.
Then, this weight assignment scheme can only distinguish the inter-node connections from the self-loop connections.
This variant is similar to GraphSAGE, which forces the neighbor projection matrix to be the multiple of residual projection matrix, and it is denoted as HA-GAT(O).

(2) When the gradient scaling factor $\lambda$ is set to a very small value, e.g., $\lambda=1e-10$, the gradients to LAPs will also be very small.
Then, each attention coefficient $w_{ij}^{(l)}$ is approximately 1.
Our HA-GAT will then degenerate to a GCN variant with Neighbor Norm.
This variant is denoted as HA-GAT(Z).

(3) BM-GCN \cite{BM-GCN} attempts to directly extract the edge heterophily according to the node labels and leverages a two-staged learning strategy.
To demonstrate the superiority of our design, we set the category dimension $t$ of our HA-GAT to the number of classes $C$, and employ the same two-staged learning strategy as \cite{BM-GCN}.
This variant is denoted as HA-GAT(T).

\subsection{Complexity Analysis}
Our HA-GAT has two backbone networks: 1) 2-layered GCN as an explorer network; 2) the classification network.

\noindent \textbf{Time Complexity.}
For explorer network, the time complexity of each layer is $O(|N|d_{in} d_{out}$ $+|E| d_{out})$, i.e., the time complexity of GCN layer.
For the classification network, in each layer, the complexity of calculating edge weight is $O(|E|t^2)$, the complexity of message aggregation step is still $O(|N|d_{in} d_{out} + |E| d_{out})$.
Thus, the total time complexity is $O(|N|d_{in} d_{out} + |E| d_{out} + |E|t^2)$.
Note that $t$ represents the number of underlying node categories, which is usually set to a relatively small number.
We analyze the impacts of $t$ in \cref{subsection-LDE}.
In practice, we set $t=3$ for all experiments in \cref{table-supervised,table-semi-supervised}.

\noindent \textbf{Memory Complexity.}
For explorer network, the memory complexity of each layer is $O(d_{in} d_{out})$, i.e., the memory complexity of GCN layer.
In the classification network, the parameters are $\boldsymbol\omega$, $\omega_{sl}$, and $\Theta$. Compared with a GCN layer, a classification network layer only introduces extra $t^2 + 1$ parameters to get the weight, so the complexity is $O(d_{in} d_{out} + t^2)$, while $t$ is a very small number, as stated above.

Overall, for an L-layered HA-GAT, we utilize a 2-layered GCN as an explorer network and an L-layered classification network. The final time complexity and memory complexity are the addition of the corresponding complexities of these (2+L) layers. Considering that $t$ is a relatively small number, e.g., $ t=3$ in our settings, the time and memory complexities of L-layered HA-GAT are approximated to those of an (L+2)-layered GCN.

\begin{table*}[t]
  \scriptsize
  \centering
  \caption{Datasets.}
  \begin{tabular}{c|ccccc|ccc}
    \toprule
                                  & Actor     & Chameleon & Squirrel  & Texas     & Cornell   & Cora      & CiteSeer  & PubMed    \\
    \midrule
      Classes                     & 5         & 5         &5          & 5         & 5         & 7         & 6         & 5         \\
      Features                    & 932       & 2325      & 2089      & 1703      & 1703      & 1433      & 3703      & 500       \\
      Nodes                       & 7600      & 2277      & 5201      & 183       & 183       & 2708      & 3327      & 19717     \\
      Edges                       & 26659     & 31371     & 198353    & 279       & 277       & 5278      & 4552      & 44324     \\
      $\mathcal{H(\mathcal{G})}$  & 0.158     & 0.248     & 0.218     & 0.087     & 0.306     & 0.825     & 0.706     & 0.762     \\
    \bottomrule
  \end{tabular}
  \label{table-statistics}
\end{table*}

\section{Evaluations}
In this section, we evaluate and analyze the effectiveness of our HA-GAT, which contains a local distribution exploration and a heterophily-aware attention scheme, on eight node classification datasets in both the supervised and semi-supervised settings, by seeking an interpretation from their mechanism.

\begin{table*}[t]
  \scriptsize
  \centering
  \caption{Results of supervised node classification: Mean accuracy (\%) ± standard deviation.
  M denotes that it occurs an out-of-memory error.
  Note that since HA-GAT(L) leverages the prior label information, it is not involved in this comparison.}
  \setlength{\tabcolsep}{4pt}
  \begin{tabular}{c|ccccc|ccc}
    \toprule
      Methods           & Actor                 & Chameleon             & Squirrel              & Texas                 & Cornell               & Cora                  & CiteSeer              & PubMed        \\
    \midrule
      MLP               & 40.26{\tiny±1.08}     & 48.94{\tiny±1.86}     & 31.43{\tiny±1.25}     & 89.29{\tiny±5.43}     & 89.07{\tiny±4.20}     & 75.97{\tiny±1.71}     & 76.01{\tiny±1.33}     & 86.78{\tiny±0.44} \\
    \midrule
      GraphSAGE         & 37.78{\tiny±1.17}     & 67.89{\tiny±1.76}     & 54.19{\tiny±1.17}     & 84.20{\tiny±4.92}     & 81.96{\tiny±7.39}     & 87.73{\tiny±1.29}     & 80.11{\tiny±1.50}     & 89.17{\tiny±0.52} \\
      GCN               & 33.31{\tiny±1.09}     & 67.96{\tiny±2.13}     & 53.85{\tiny±1.25}     & 77.00{\tiny±6.82}     & 77.54{\tiny±4.21}     & 87.17{\tiny±1.23}     & 80.80{\tiny±1.15}     & 87.71{\tiny±0.42} \\
      GAT               & 34.44{\tiny±1.06}     & 65.89{\tiny±2.60}     & 51.45{\tiny±2.59}     & 78.03{\tiny±4.31}     & 78.63{\tiny±4.26}     & 87.39{\tiny±1.27}     & 80.46{\tiny±1.47}     & 87.15{\tiny±0.42} \\
      GATv2             & 34.94{\tiny±1.21}     & 66.93{\tiny±2.25}     & 50.27{\tiny±1.61}     & 79.29{\tiny±4.03}     & 80.33{\tiny±3.33}     & \textbf{88.43{\tiny±1.31}}& 80.52{\tiny±1.18} & 88.06{\tiny±0.40} \\
    \midrule
      Geom-GCN          & 31.81{\tiny±0.24}     & 61.06{\tiny±0.49}     & 38.28{\tiny±0.27}     & 55.59{\tiny±1.59}     & 58.56{\tiny±1.77}     & 85.09{\tiny±1.44}     & 77.78{\tiny±1.62}     & 88.59{\tiny±0.47} \\
      JKNet-GCN         & 34.41{\tiny±1.30}     & 65.58{\tiny±2.03}     & 50.11{\tiny±3.66}     & 79.62{\tiny±3.94}     & 77.16{\tiny±4.21}     & 86.33{\tiny±1.53}     & 79.05{\tiny±1.57}     & 88.37{\tiny±0.65} \\
      ChebyNet          & 39.17{\tiny±1.43}     & 62.40{\tiny±2.21}     & 43.36{\tiny±1.29}     & 83.01{\tiny±6.03}     & 84.59{\tiny±4.49}     & 87.11{\tiny±1.17}     & 80.02{\tiny±1.58}     & 88.34{\tiny±0.46} \\
      H2GCN             & 36.35{\tiny±0.76}     & 65.43{\tiny±2.27}     & 58.13{\tiny±1.59}     & 69.47{\tiny±5.13}     & 73.68{\tiny±7.67}     & 80.00{\tiny±0.70}     & 64.06{\tiny±1.52}     & 76.00{\tiny±0.44} \\
      DMP               & 37.26{\tiny±1.79}     & M                     & M                     & 81.64{\tiny±5.20}     & 87.00{\tiny±5.52}     & 84.03{\tiny±1.25}     & 74.61{\tiny±1.78}     & M                 \\
      GPR-GNN           & 39.21{\tiny±1.27}     & 67.52{\tiny±2.37}     & 52.06{\tiny±1.85}     & 89.52{\tiny±4.41}     & 92.79{\tiny±3.12}     & 88.41{\tiny±1.35}     & 80.27{\tiny±1.42}     & 89.31{\tiny±0.87} \\
      BM-GCN            & 39.56{\tiny±1.43}     & 66.93{\tiny±1.92}     & 47.32{\tiny±1.42}     & 78.14{\tiny±4.80}     & 81.97{\tiny±6.13}     & 87.34{\tiny±1.22}     & 79.65{\tiny±1.72}     & 87.00{\tiny±0.74} \\
    \midrule
      HA-GAT            & 41.96{\tiny±1.36} &\textbf{72.58{\tiny±2.08}} & \textbf{69.02{\tiny±1.50}} & \textbf{91.75{\tiny±3.52}} & \textbf{93.44{\tiny±2.84}} & 88.26{\tiny±1.24} & \textbf{81.91{\tiny±1.20}} & \textbf{89.52{\tiny±0.52}}    \\
    \midrule
      HA-GAT(G)         & 40.16{\tiny±1.20}     & 67.14{\tiny±2.40}     & 63.39{\tiny±1.31}     & 86.89{\tiny±7.68}     & 88.74{\tiny±5.69}     & 87.11{\tiny±1.45}     & 80.97{\tiny±1.35}     & 88.30{\tiny±0.61} \\
      HA-GAT(M)         & \textbf{42.16{\tiny±1.14}} & 69.22{\tiny±1.68} & 62.73{\tiny±1.49}    & 91.28{\tiny±3.22}     & 92.84{\tiny±3.52}     & 87.76{\tiny±1.40}     & 81.05{\tiny±1.01}     & 88.76{\tiny±0.54} \\
      HA-GAT(O)         & 41.73{\tiny±1.21}     & 69.31{\tiny±2.51}     & 57.40{\tiny±5.44}     & 88.19{\tiny±6.25}     & 89.72{\tiny±4.76}     & 87.80{\tiny±1.22}     & 80.28{\tiny±1.67}     & 88.42{\tiny±0.56} \\
      HA-GAT(Z)         & 33.39{\tiny±1.09}     & 67.53{\tiny±2.10}     & 53.08{\tiny±1.56}     & 78.85{\tiny±3.86}     & 77.32{\tiny±5.52}     & 88.15{\tiny±1.11}     & 80.44{\tiny±1.63}     & 87.79{\tiny±0.48} \\
      HA-GAT(T)         & 35.07{\tiny±3.09}     & 70.75{\tiny±2.02}     & 61.00{\tiny±1.89}     & 84.48{\tiny±6.51}     & 79.51{\tiny±7.08}     & 86.80{\tiny±1.54}     & 80.27{\tiny±1.58}     & 87.93{\tiny±0.53} \\
    \midrule
    \textit{HA-GAT(L)} & \textit{63.38{\tiny±5.38}} & \textit{92.26{\tiny±1.43}} & \textit{85.46{\tiny±5.52}} & \textit{91.86{\tiny±3.37}} & \textit{92.13{\tiny±2.18}} & \textit{93.26{\tiny±1.29}} & \textit{87.29{\tiny±1.83}} & \textit{96.90{\tiny±0.99}} \\
    \bottomrule
  \end{tabular}
  \label{table-supervised}
\end{table*}

\subsection{Experimental Settings}
\noindent\textbf{Datasets.}
We adopt eight node classification benchmarks with varying homophily ratios. Based on their homophily ratios, calculated using \cref{eq-homophily_ratio}, they can be categorized into five heterophilic networks and three homophilic networks. Specifically, the homophilic networks contain three citation networks, including Cora, CiteSeer and PubMed \cite{citation,planetoid}. The heterophilic networks include two Wikipedia networks (i.e., Chameleon and Squirrel), one co-occurrence network (i.e., Actor), and two WebKB webpage networks (i.e., Texas and Cornell) \cite{geom-gcn}. Their details are presented in \cref{table-statistics}.

\noindent\textbf{Baselines.}
We compare our method with 12 state-of-the-art methods.
(1) One graph-agnostic method: Multi-layer Perception Machine (MLP);
(2) Four Standard GNNs: GCN \cite{gcn}, GAT \cite{gat}, GATv2 \cite{gatv2} and GraphSAGE \cite{graphsage};
(3) Seven heterophilic GNNs: JKNet-GCN \cite{jk-net}, ChebyNet \cite{gcn2}, Geom-GCN \cite{geom-gcn}, H2GCN \cite{H2GCN}, GPR-GNN \cite{gpr-gnn}, DMP \cite{DMP} and BM-GCN \cite{BM-GCN}.

\noindent\textbf{Setups.}
For the supervised node classification tasks, we follow the strategy in \cite{gpr-gnn}, which randomly splits 60\%/20\%/20\% nodes into the training/validation/testing sets.
For the semi-supervised setting, we adopt the public splits \cite{planetoid} for Cora, CiteSeer, and Pubmed, and randomly split 10\%/10\%/80\% nodes as training/validation/testing samples on other heterophilic datasets.
Every experiment is run 100 times with different random splits and initializations.
In our evaluations, the category dimension $t$ is set to 3 and the gradient scaling factor is varied in $\left\{0.1, 1.0, 10\right\}$.
A two-layered HA-GAT with a 64-neuron hidden layer is employed for both the supervised and semi-supervised settings.
The cross-entropy loss is utilized as the loss function, and Adam \cite{adam} optimizer is employed.
The learning rate, weight decay rate, and dropout rate are obtained by the grid-search strategy.
Note that for fair comparisons, we utilize the same grid-search strategy for these existing GNNs.

\begin{figure}[t]
  \centering
  \begin{subfigure}[b]{0.35\columnwidth}
    \includegraphics[width=\linewidth]{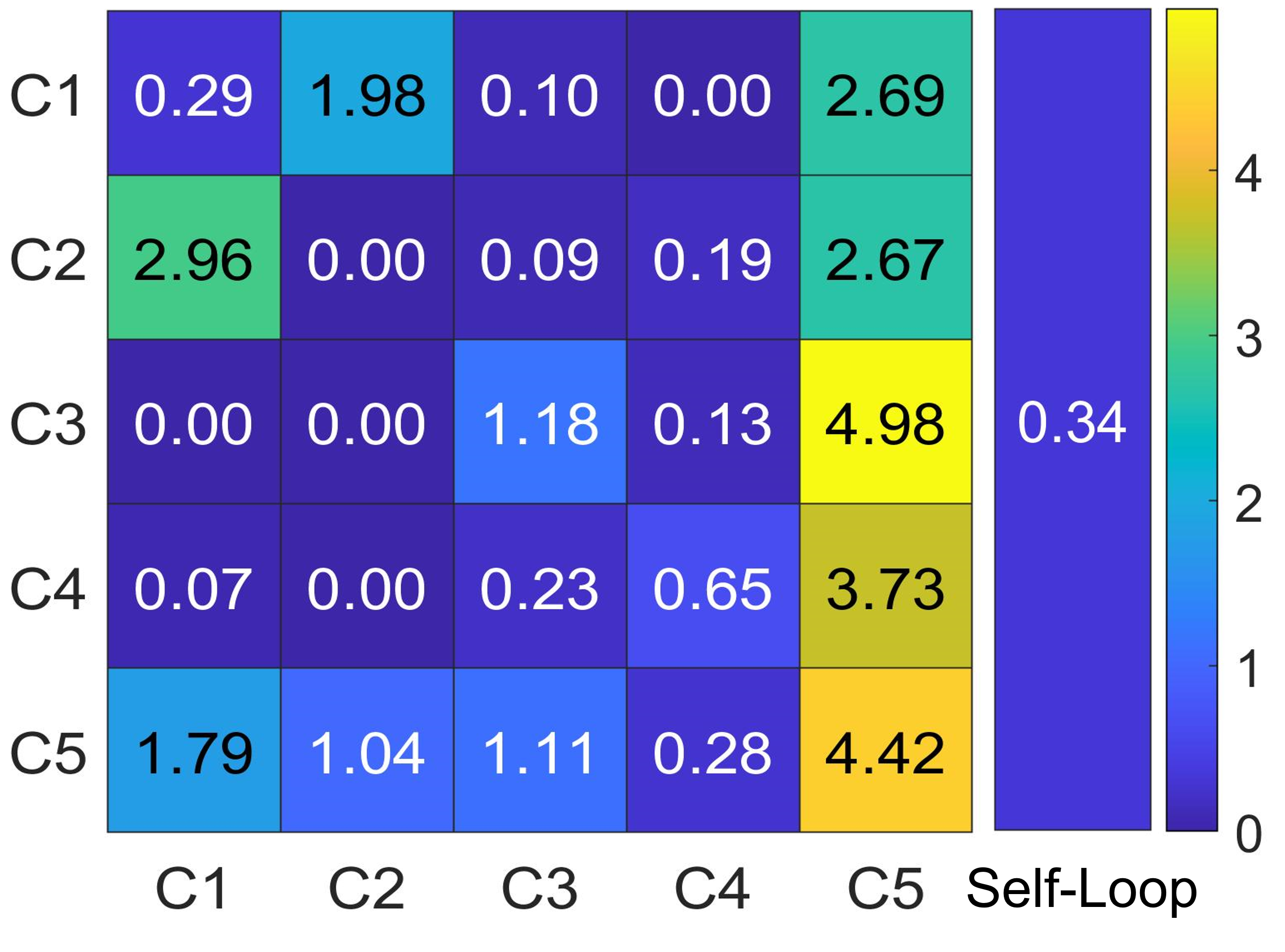}
    \caption{LAP in layer 1.}
    \label{fig-label-attention-ew1}
  \end{subfigure}
  \hspace{1cm}
  \begin{subfigure}[b]{0.35\columnwidth}
    \includegraphics[width=\linewidth]{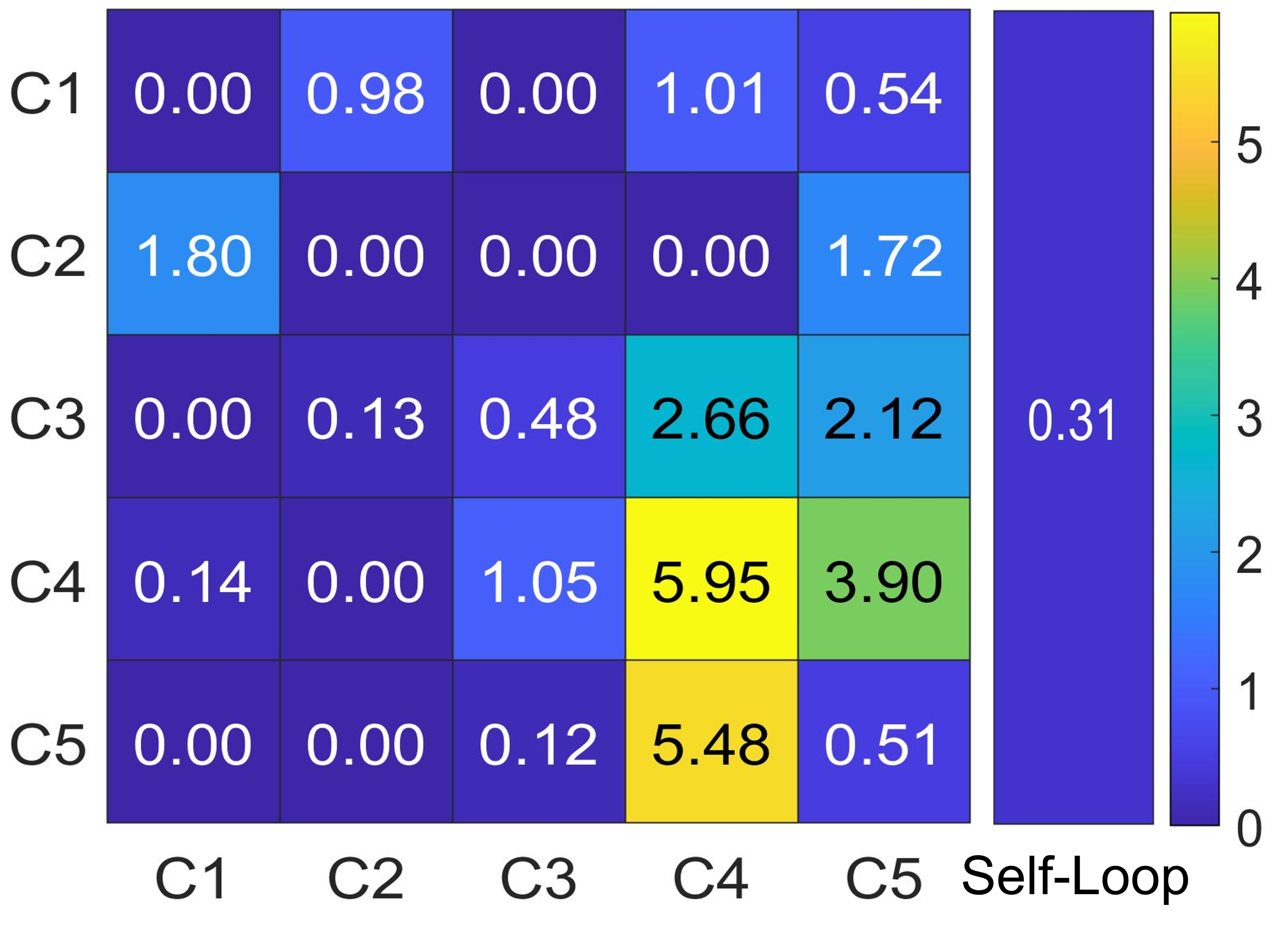}
    \caption{LAP in layer 2.}
    \label{fig-label-attention-ew2}
  \end{subfigure}
  \caption{Visualization of the LAPs for a 2-layered HA-GAT(L) on Chameleon.}
  \label{fig-label-attention-ew}
\end{figure}

\begin{table*}[]
  \scriptsize
  \centering
  \caption{Results of semi-supervised node classification: Mean accuracy (\%) ± standard deviation.}
  \setlength{\tabcolsep}{4pt}
  \begin{tabular}{c|ccccc|ccc}
    \toprule
      Methods   & Actor                 & Chameleon             & Squirrel              & Texas                 & Cornell               & Cora                  & CiteSeer              & PubMed        \\
    \midrule
      MLP       & 33.47{\tiny±0.47}     & 37.13{\tiny±1.20}     & 25.69{\tiny±0.87}     & 65.81{\tiny±4.55}     & 70.06{\tiny±5.12}     & 60.07{\tiny±1.30}     & 60.82{\tiny±1.01}     & 73.12{\tiny±0.48} \\
      GCN       & 26.06{\tiny±1.21}     & 52.32{\tiny±1.42}     & 31.26{\tiny±1.83}     & 56.35{\tiny±4.95}     & 49.93{\tiny±5.29}     & 81.69{\tiny±0.67}     & 71.58{\tiny±0.55}     & 79.15{\tiny±0.49} \\
      GAT       & 27.37{\tiny±1.04}     & 52.85{\tiny±1.89}     & 31.27{\tiny±1.72}     & 53.78{\tiny±4.14}     & 53.85{\tiny±7.79}     & 83.04{\tiny±0.73}     & 71.48{\tiny±0.69}     & 78.50{\tiny±0.38} \\
      GPR-GNN   & 32.51{\tiny±1.05}     & 47.13{\tiny±4.10}     & 33.11{\tiny±1.81}     & 65.54{\tiny±6.85}     & 65.54{\tiny±6.85}     & \textbf{83.43{\tiny±0.69}} & 71.41{\tiny±0.80} & 79.21{\tiny±0.53} \\
      BM-GCN    & 30.84{\tiny±0.90}     & 54.13{\tiny±1.41}     & 33.94{\tiny±1.34}     & 64.86{\tiny±7.00}     & 54.86{\tiny±9.32}     & 81.01{\tiny±0.51}     & 69.34{\tiny±1.12}     & 76.50{\tiny±0.62} \\
    \midrule
      HA-GAT    & \textbf{35.14{\tiny±0.68}} & \textbf{58.84{\tiny±1.84}} & \textbf{45.67{\tiny±0.97}} & \textbf{75.47{\tiny±5.32}} & \textbf{73.71{\tiny±5.66}} & \textbf{83.43{\tiny±0.43}} & \textbf{72.51{\tiny±0.62}} & \textbf{79.68{\tiny±0.34}} \\
    \bottomrule
  \end{tabular}
  \label{table-semi-supervised}
\end{table*}

\subsection{Effectiveness of Heterophily-Aware Attention Scheme}
The proposed heterophily-aware attention scheme attempts to find an effective LAP to help nodes to extract appropriate information from both similar and dissimilar neighbors.
As can be observed in \cref{table-supervised}, by exploiting the prior edge heterophily to identify edges according to node labels, HA-GAT(L) achieves superior results on all the datasets, especially on the heterophilic graphs.
It indicates that even on the heterophilic graphs, where the nodes usually connect with other dissimilar nodes, there still exists an effective LAP which can enable the nodes to acquire appropriate information from distinct neighbors.
Note that since Texas and Cornell are quite small, their performances are vulnerable to the topological bias.
\cref{fig-label-attention-ew} visualizes the learned LAPs of HA-GAT(L) on Chameleon, where the specific weight assignments from each type of nodes to all their neighbors can be observed.
For example, in the first layer, the edges coming into the nodes of C5 have large weights, which suggests that the nodes of C5 may possess salient characteristics after the aggregation.
Then, the classification tasks may benefit accordingly.
(Note that the nodes of C5 represent the web pages with the most significant averaged traffic.)

\subsection{Main Results of HA-GAT}

When the prior edge heterophily is not given, our HA-GAT excavates effective local distributions as a substitute for the underlying heterophily.
Here, we demonstrate the effectiveness of the proposed HA-GAT for both the supervised and semi-supervised node classification on real-world graphs with different homophily ratios.

\begin{figure}[t]
  \centering
  \begin{subfigure}[b]{0.35\columnwidth}
    \includegraphics[width=\linewidth]{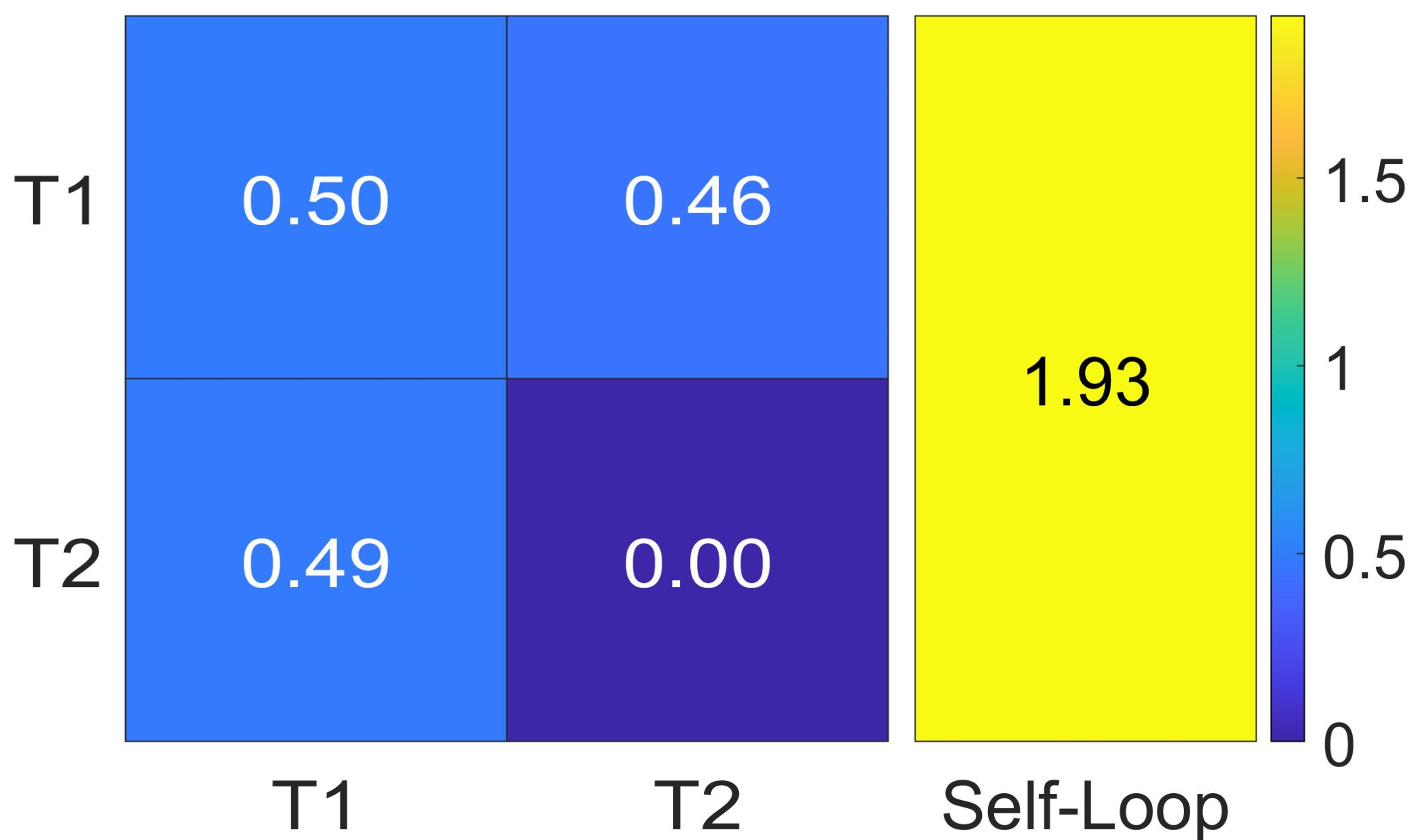}
    \caption{LAP of the first layer in HA-GAT.}
    \label{fig-actor-gcn-ew1}
  \end{subfigure}
  \hspace{1cm}
  \begin{subfigure}[b]{0.35\columnwidth}
    \includegraphics[width=\linewidth]{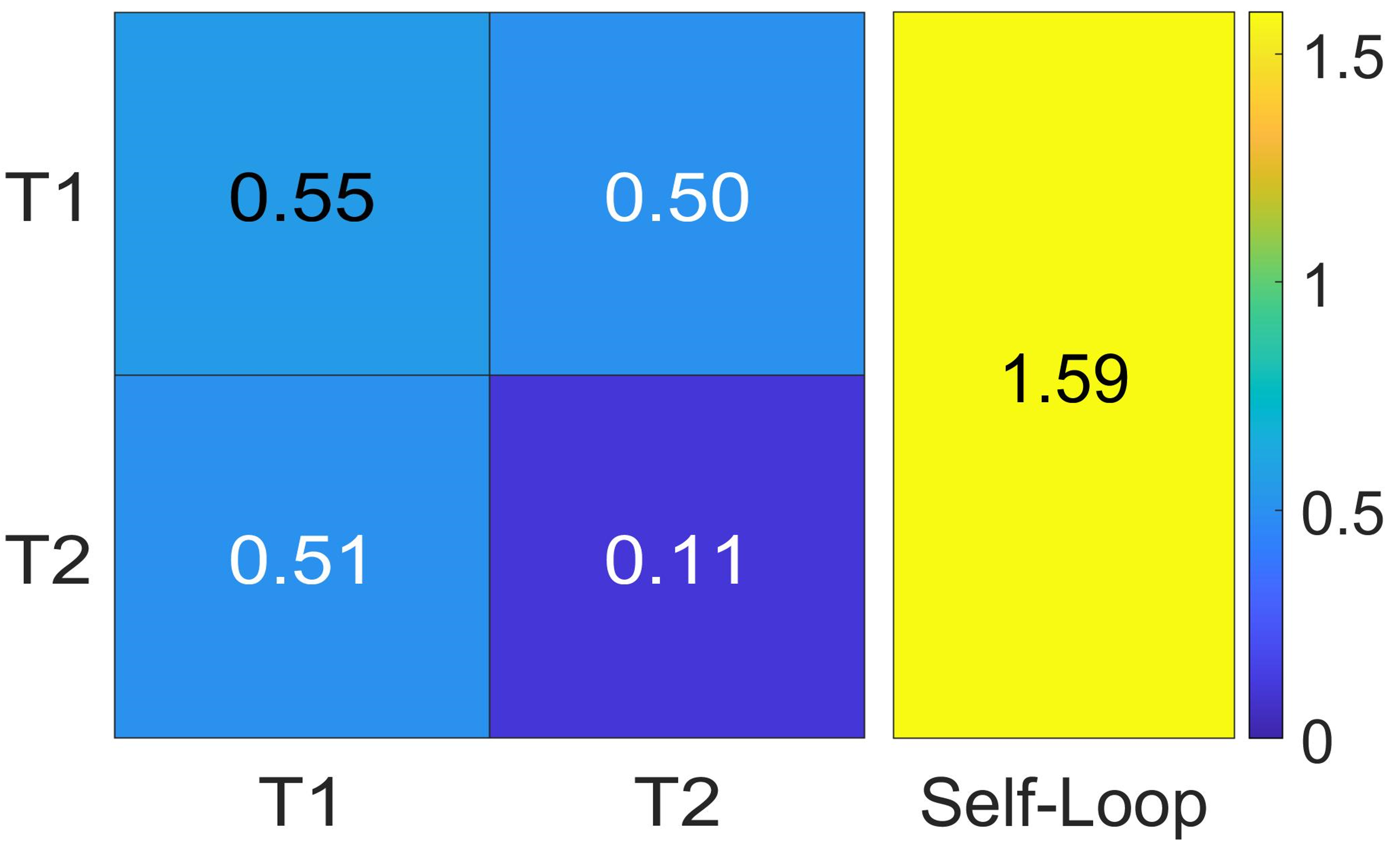}
    \caption{LAP of the first layer in HA-GAT(M).}
    \label{fig-actor-mlp-ew1}
  \end{subfigure}
  \caption{Visualization of the LAPs for a 2-layered HA-GAT with different explorer networks on Actor.}
  \label{fig-actor-explorer-network}
\end{figure}

\noindent\textbf{Supervised Node Classification.}
Recent studies \cite{ma2021homophily,wang2024understanding} have proved that the standard GNNs, such as GCN, can also achieve decent performances on the heterophilic graphs, where the node local distributions are distinguishable, e.g., on the Chameleon and Squirrel datasets.
As can be observed in \cref{table-supervised}, all the GNNs outperform MLP on these two datasets.
Our HA-GAT, which excavates effective local distributions and leverages a heterophily-aware attention scheme, can further improve the weight assignment scheme, and achieve superior performances.
For example, GCN improves 22.43\% accuracy on Squirrel over MLP, because of the distinguishable local distributions, and obtains a comparable performance to the existing heterophilic GNNs.
Our HA-GAT further achieves an up to 15.17\% gain over GCN due to our heterophily-aware attention scheme, by fully considering the local distributions.
On the Actor, Cornell, and Texas datasets, where the node local distributions are indistinguishable \cite{ma2021homophily}, the accuracies of standard GNNs, e.g., GCN and GAT, are lower than those of MLP.
The heterophilic GNNs, which emphasize ego-representations, e.g., GPR-GNN, can achieve excellent performances.
However, as can be observed, few methods perform better than MLP on Actor.
Since our attention scheme leverages an adaptive weight to model the importance of the self-loop connections and excavates useful information among neighborhoods, our HA-GAT outperforms the others on these datasets.
As observed in \cref{fig-actor-gcn-ew1}, both the edges and self-loops are considered.
Besides, our HA-GAT(Z) and HA-GAT(O), which are the equivalents of certain variants of GCN and GraphSAGE, achieve similar performances compared to their corresponding counterparts.
Overall, the proposed HA-GAT achieves state-of-the-art performances on all the datasets, hence it is applicable to the graphs with different homophily ratios.

\noindent\textbf{Semi-Supervised Node Classification.}
Under the semi-supervised settings, as can be observed from \cref{table-semi-supervised}, all the results decrease significantly compared to the results under the supervised settings.
However, the relative performances of the baselines on different graphs are similar to these under the supervised settings.
For example, MLP outperforms GCN on Actor, while GCN achieves better performances on other datasets.
An exception is BM-GCN, which directly estimates the edge heterophily according to the node labels and achieves relatively worse performances on three homophilic datasets than other GNN baselines.
It suggests that estimating the label heterophily is not applicable under the semi-supervised settings, where only the minority of nodes are labeled (e.g., CiteSeer only provides 120/500 labeled nodes for training and validation).
On the contrary, the proposed HA-GAT can still excavate effective local distributions to represent the underlying heterophily, and achieves state-of-the-art results on all the datasets under the semi-supervised settings, which verifies the superiority of our method.

\begin{figure}[t]
  \centering
  \begin{subfigure}[b]{0.22\columnwidth}
    \includegraphics[trim=0cm 0cm 0.3cm 0cm, clip, width=\textwidth]{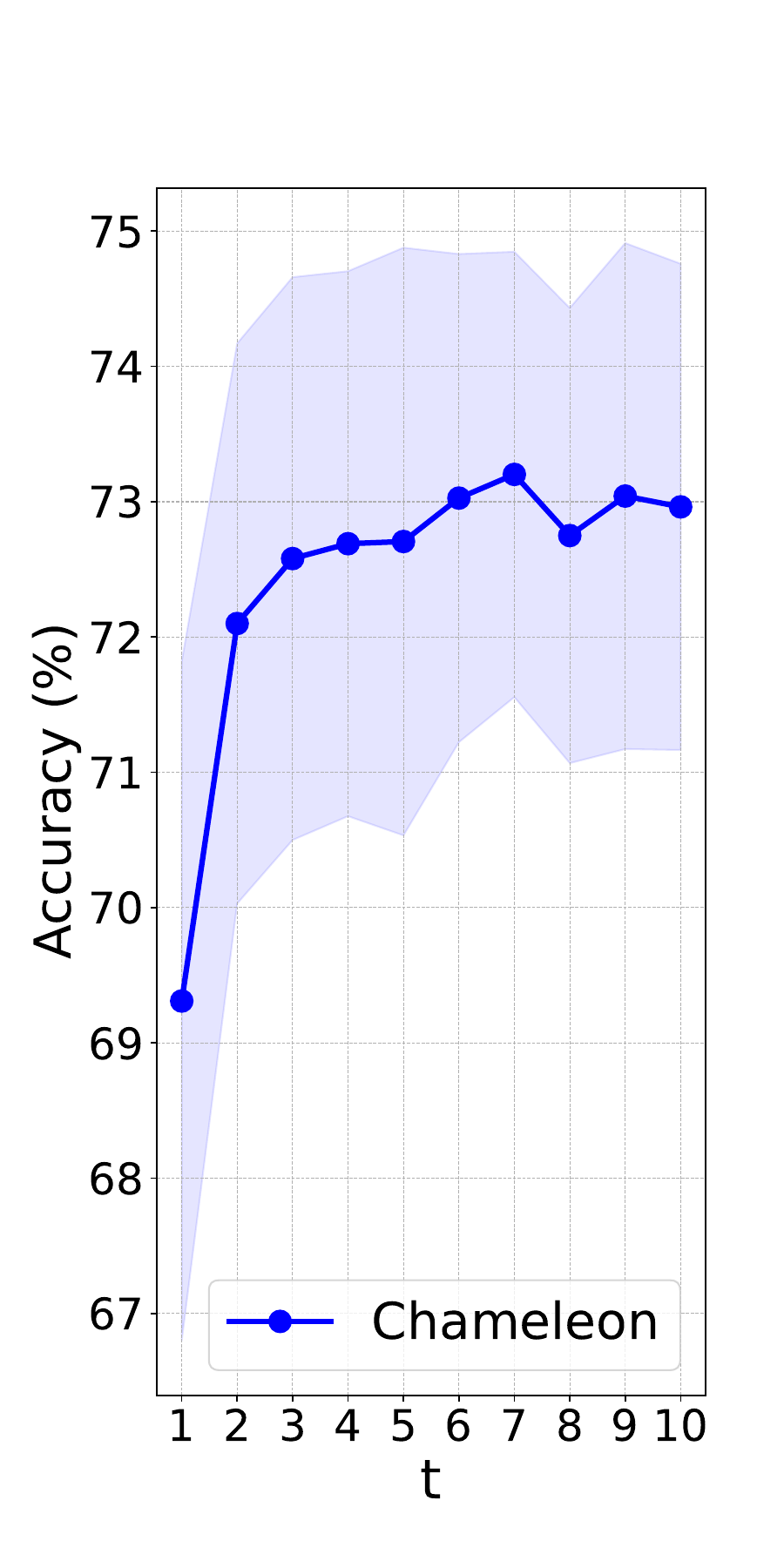}
    \caption{Accuracy with diverse $t$.}
    \label{fig4:a}
  \end{subfigure}
  \begin{minipage}[b]{0.77\textwidth}
    \begin{subfigure}[b]{0.33\linewidth}
      \includegraphics[width=\linewidth]{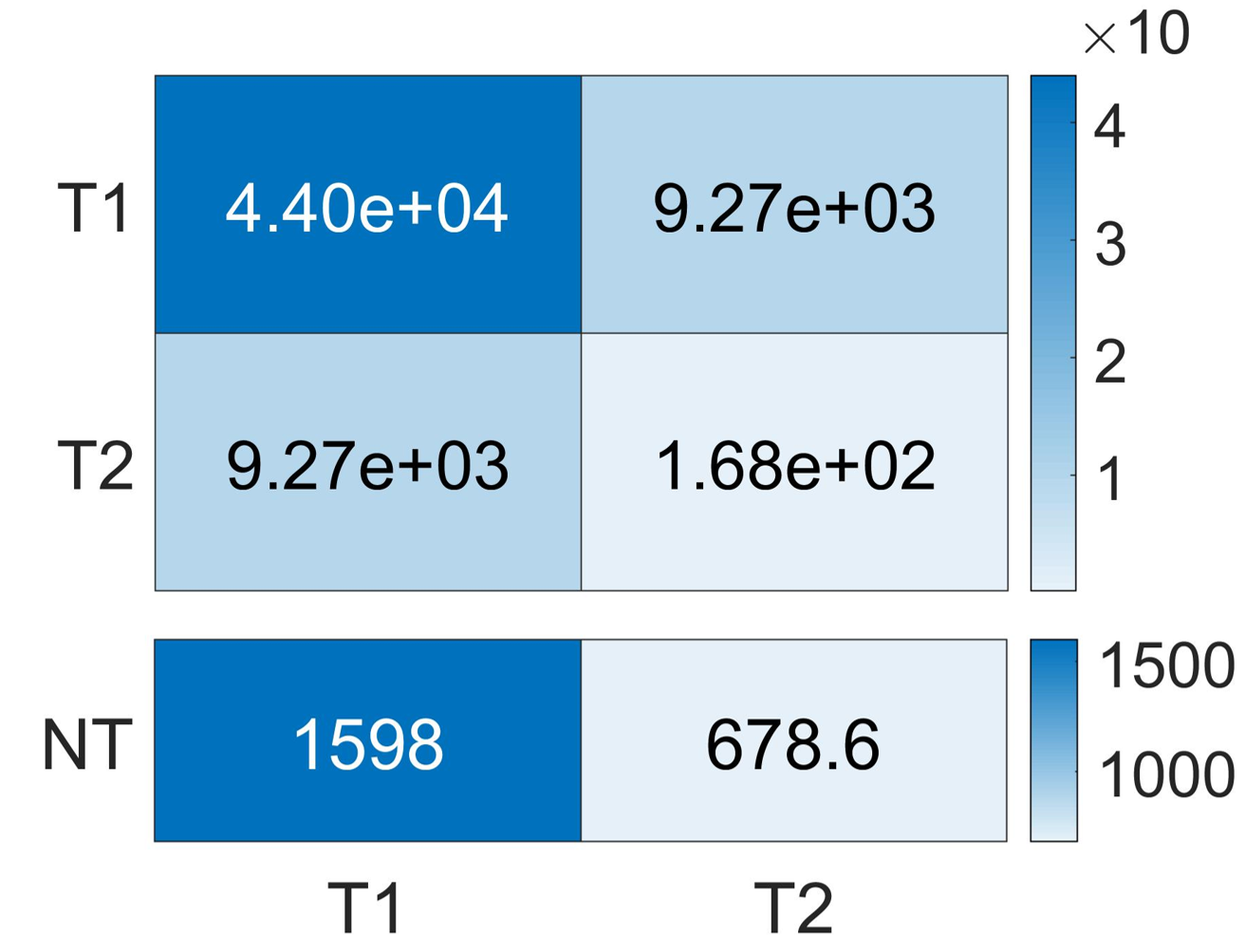}
      \caption{$M$ and $N_T$ ($t=2$).}
      \label{fig4:b}
    \end{subfigure}\hfill
    \begin{subfigure}[b]{0.33\linewidth}
      \includegraphics[width=\linewidth]{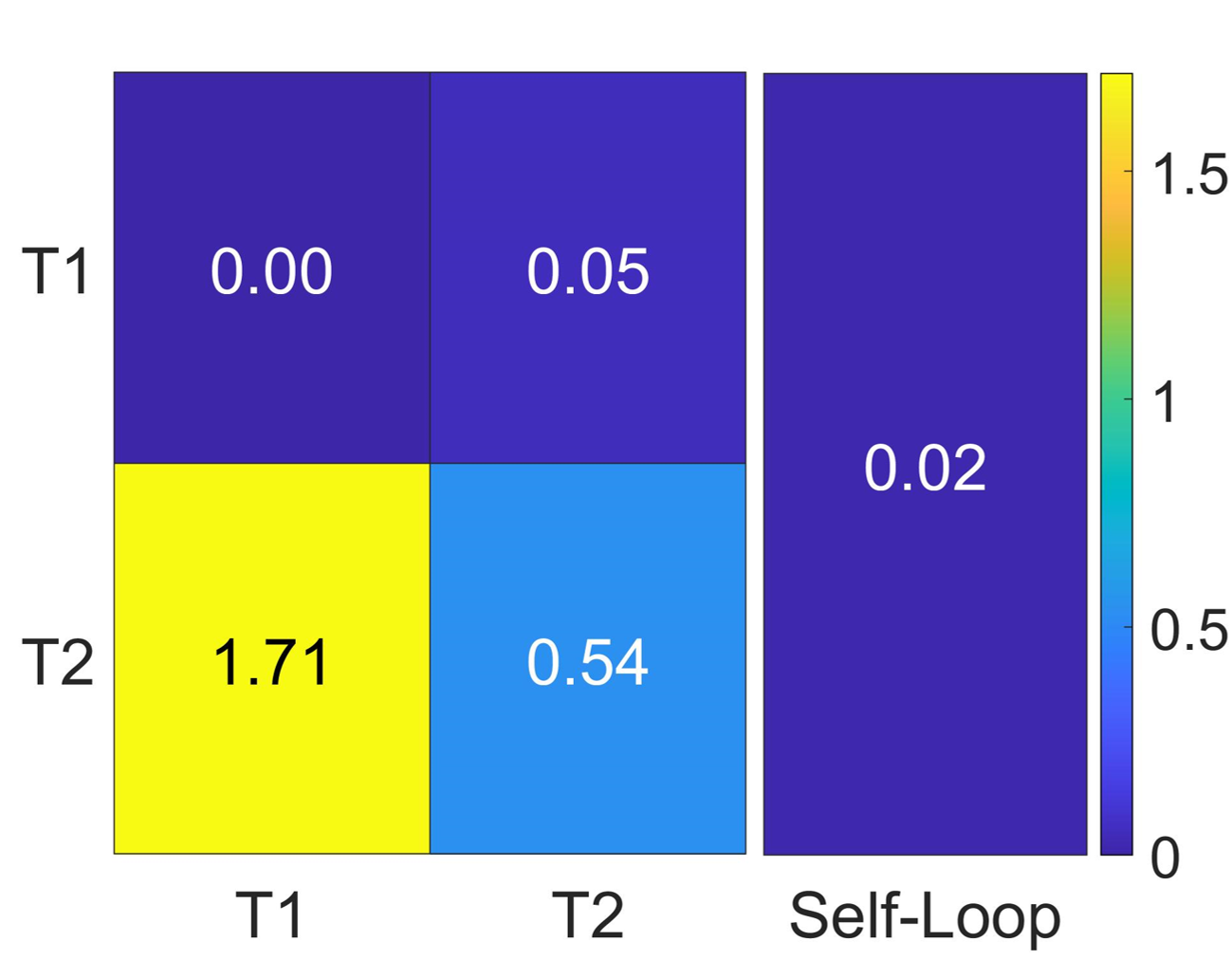}
      \caption{LAP in Layer $1$ ($t=2$).}
      \label{fig4:c}
    \end{subfigure}\hfill
    \begin{subfigure}[b]{0.33\linewidth}
      \includegraphics[width=\linewidth]{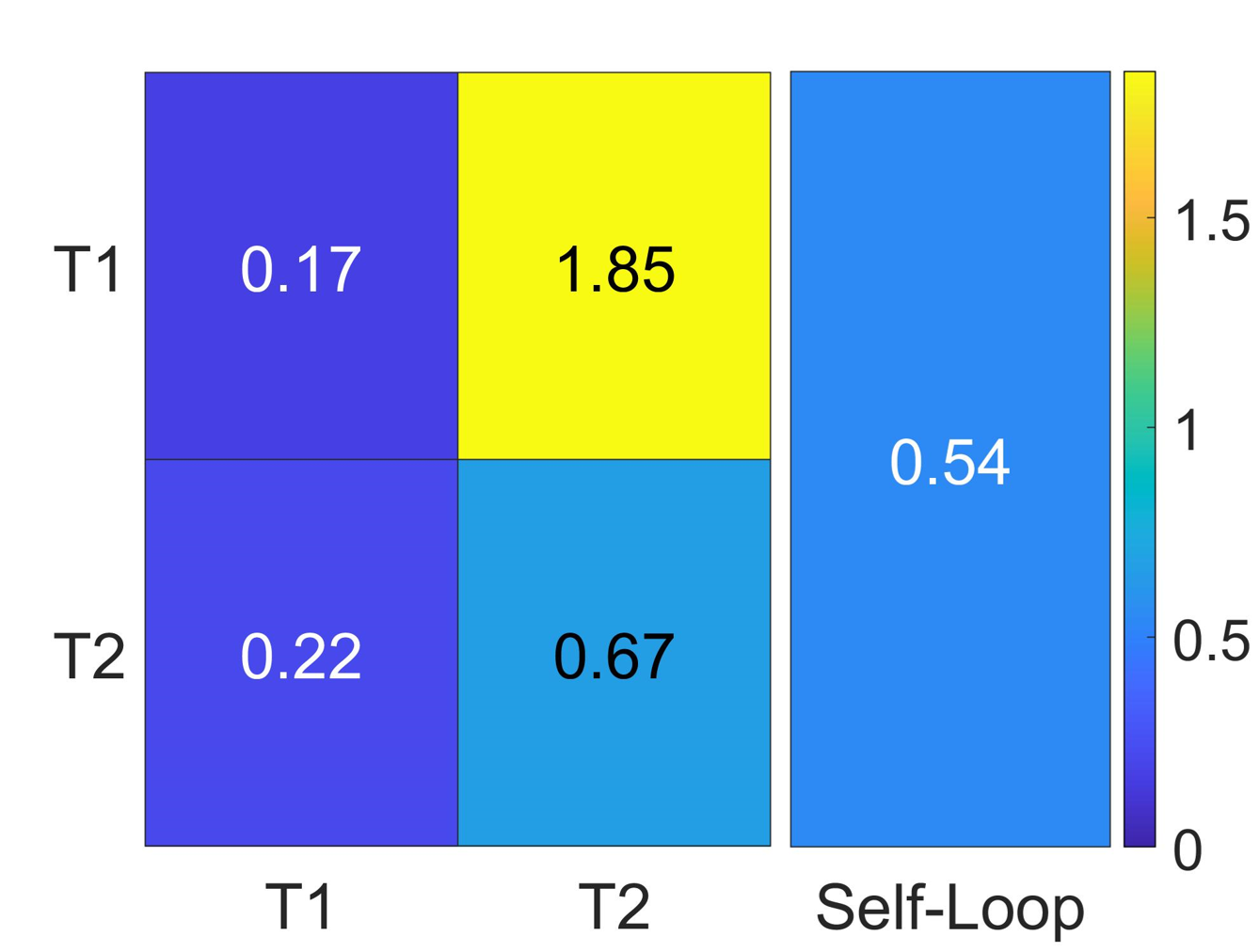}
      \caption{LAP in Layer $2$ ($t=2$).}
      \label{fig4:d}
    \end{subfigure}
    
    \vspace{1em} 
    \begin{subfigure}[b]{0.33\linewidth}
      \includegraphics[trim=0.3cm 0cm 0cm 0cm, clip, width=\linewidth]{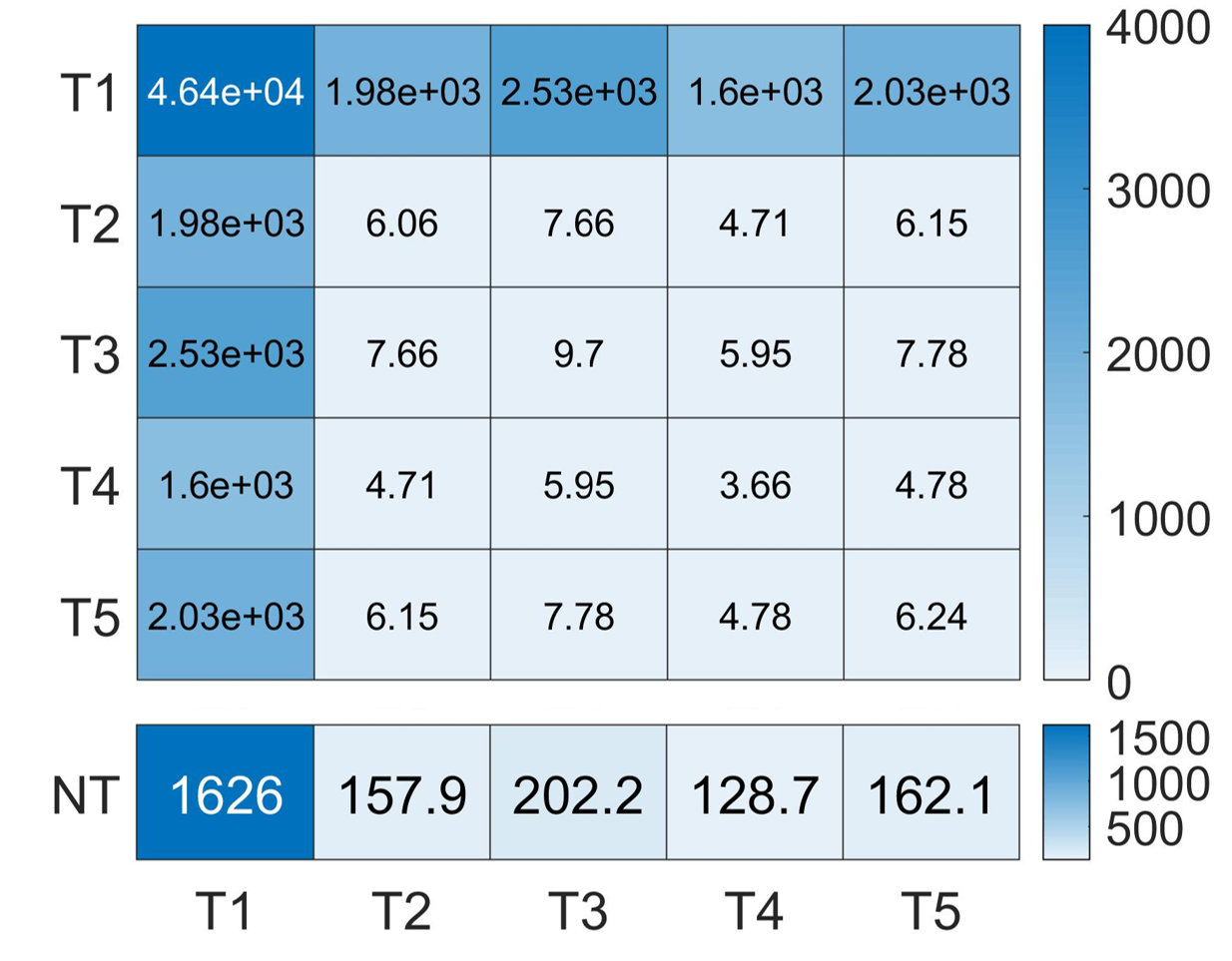}
      \caption{$M$ and $N_T$ ($t=5$).}
      \label{fig4:e}
    \end{subfigure}\hfill
    \begin{subfigure}[b]{0.33\linewidth}
      \includegraphics[width=\linewidth]{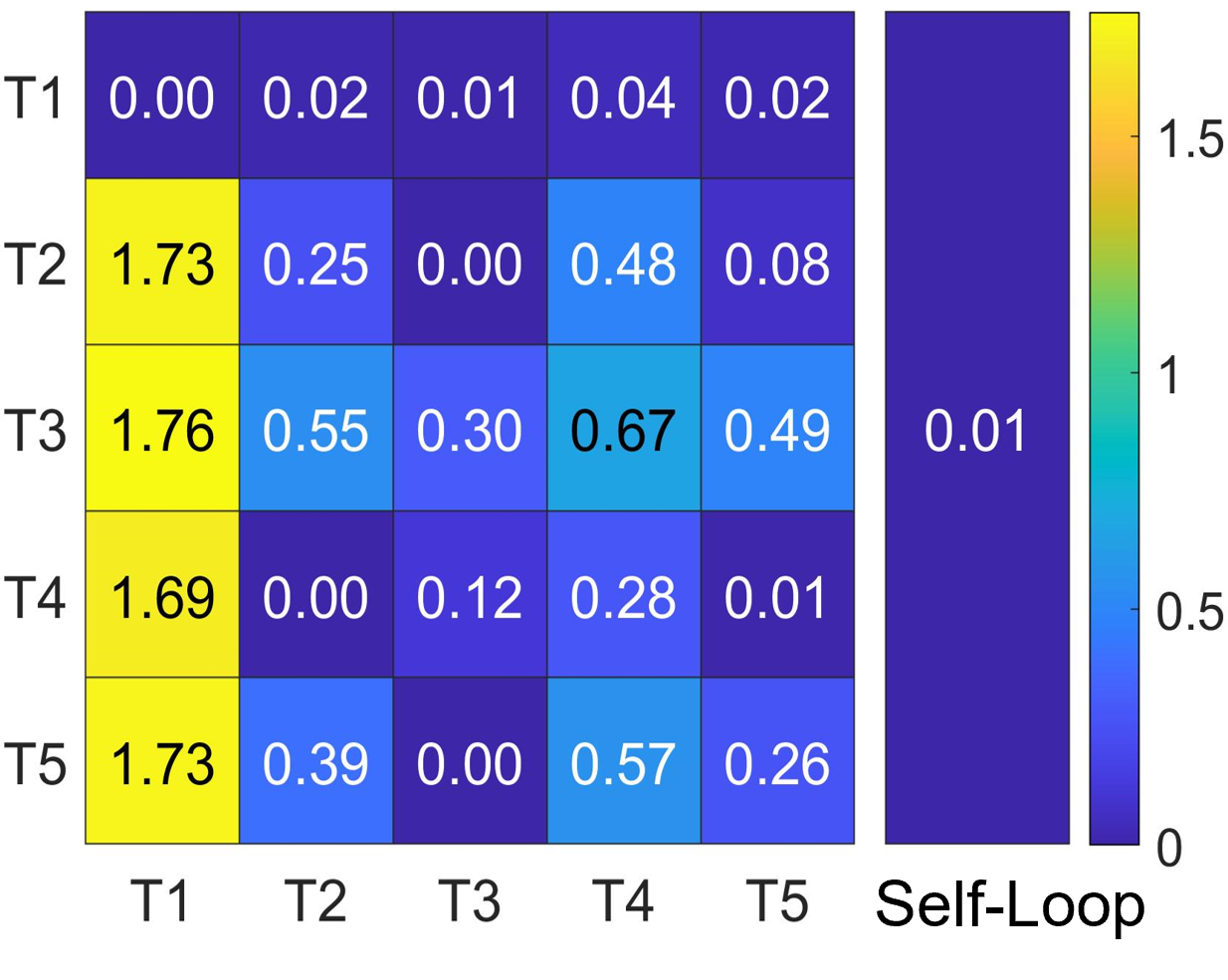}
      \caption{LAP in Layer $1$ ($t=5$).}
      \label{fig4:f}
    \end{subfigure}\hfill
    \begin{subfigure}[b]{0.33\linewidth}
      \includegraphics[width=\linewidth]{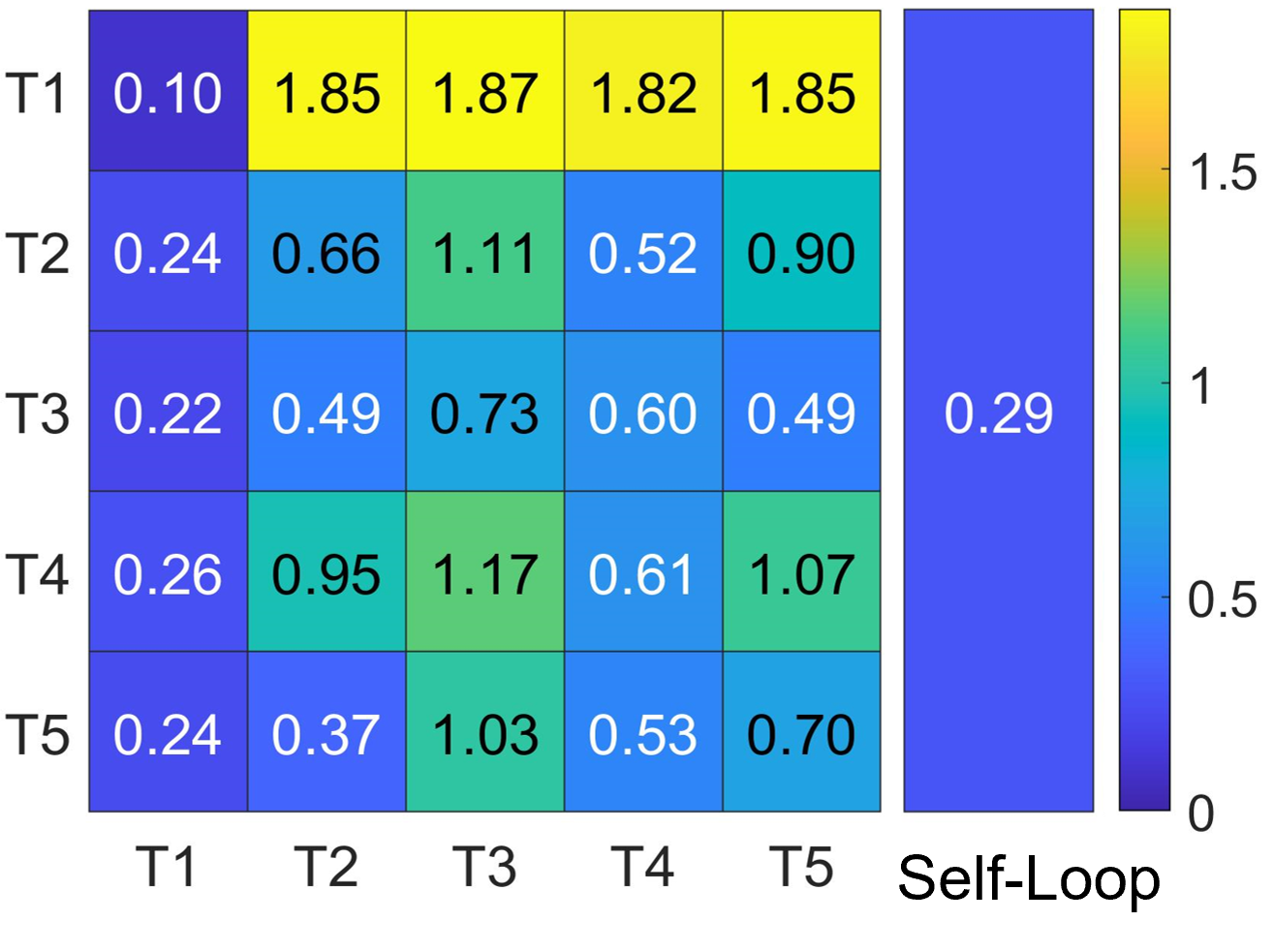}
      \caption{LAP in Layer $2$ ($t=5$).}
      \label{fig4:g}
    \end{subfigure}
  \end{minipage}
  \caption{Results of 2-layered HA-GATs on the Chameleon dataset. 
    (a) shows the accuracies of HA-GAT with different category dimension $t$. 
    (b) is the heatmaps of the overall heterophily preference (Top) $\boldsymbol{M}=\sum_{e_{ij}}\mathbf{m_{ij}}$ and the overall node categories (Bottom) $N_T=\sum_{v_i}s_i$ with $t=3$.
    (c) and (d) are the LAPs of HA-GAT.
    (e)\textasciitilde(g) are the heatmaps of HA-GAT with $t=5$.
    }
  \label{fig-LDE}
\end{figure}

\subsection{Analysis of the Local Distribution Exploration}
\label{subsection-LDE}
Our HA-GAT excavates effective local distributions from the local features and topology information, to represent the underlying heterophily.
The local distribution exploration is thoroughly examined and analyzed in this subsection.

\noindent\textbf{Effectiveness of the Local Distribution Exploration.}
As can be observed in \cref{table-supervised}, on the Chameleon and Squirrel datasets, our HA-GAT performs much better than HA-GAT(M), i.e., a HA-GAT variant with MLP being the explorer network.
Note that on the Squirrel dataset, our HA-GAT achieves a 6.29\% gain, which demonstrates the necessity of local distribution exploration.
On heterophilic graphs with indistinguishable local label distributions, e.g., Actor, Texas, and Cornell, HA-GAT and HA-GAT(M) achieve approximately identical performances. 
These results suggest that the explorer networks tend to excavate similar information by employing GCN and MLP.
Besides, as can be observed in \cref{fig-actor-explorer-network}, the weight for the self-loop connections is dominant in both HA-GAT and HA-GAT(M) on Actor. Meanwhile, the edge heterophily possesses limited effects, which can further interpret their approximately identical performances.
In addition, we also leverage two variants of our HA-GAT, i.e., HA-GAT(G) and HA-GAT(T) for more comparisons.
In general, our HA-GAT outperforms these two variants on all the datasets, which further verifies our superiority.

\noindent\textbf{Impacts of the Dimension t.}
The explorer network attempts to model the local distribution among $t$ categories.
\cref{fig4:a} gives the accuracies of our HA-GAT on Chameleon with different category dimensions $t$.
As can be observed, the performance of HA-GAT is substantially improved by increasing $t$ from 1 to 2, which indicates that assigning nodes to only two underlying categories can already effectively model the underlying heterophily.
When $t$ is greater than 2, the accuracy still increases, which suggests that our local distribution exploration can provide a greater representation ability with a larger category dimension.

\noindent\textbf{Going Deeper to the Mechanism.}
To better understand the mechanism of our local distribution exploration, we employ $\boldsymbol{M}=\sum_{e_{ij}}\mathbf{m_{ij}}$ (i.e., the overall heterophily preference matrix) and $N_T=\sum_{v_i}s_i$ (i.e., the overall node categories), to analyze the overall distributions of the heterophily information captured by our HA-GAT.
According to the learned local distributions, the nodes are softly classified.
As shown in \cref{fig4:b}, when $t=2$, nodes are classified into two underlying categories, T1 (with a score of 1598) and T2 (with a score of 678).
According to $M$, the majority of the edges connect different nodes within T1, while others mainly connect a node in T1 with a node in T2.
Note that only the minority of the edges connect different nodes within T2.
\textit{This phenomenon indicates that HA-GAT can effectively capture the underlying heterophily information, i.e., it sparsely selects nodes (as the nodes in T2) which are distinctive from the others in the local neighborhood.}
Then, according to the learned LAPs, which are shown in \cref{fig4:c,fig4:d}, the weight of the edge (which connects distinctive nodes), is well modeled.
When $t$ is increased to a larger number, a similar phenomenon can be observed, as shown in \cref{fig4:e}, i.e., the edges, which connect the nodes in the underlying categories 2 to 5, are seldom.
By combining the underlying categories 2 to 5, we can generate similar LAPs and distribution, compared to these when $t=2$.
\textit{The results indicate that the improvement from $t=2$ to $t=5$ is achieved because the network attempts to further classify the selected nodes into four underlying categories.}
Then, the performances can further benefit from increasing $t$, as shown in \cref{fig4:a}.

\subsection{Visualization of the Learned Weights}
\label{appendix-weight}

\begin{figure}[t]
  \centering
  \begin{subfigure}[b]{0.45\columnwidth}
    \includegraphics[width=\linewidth]{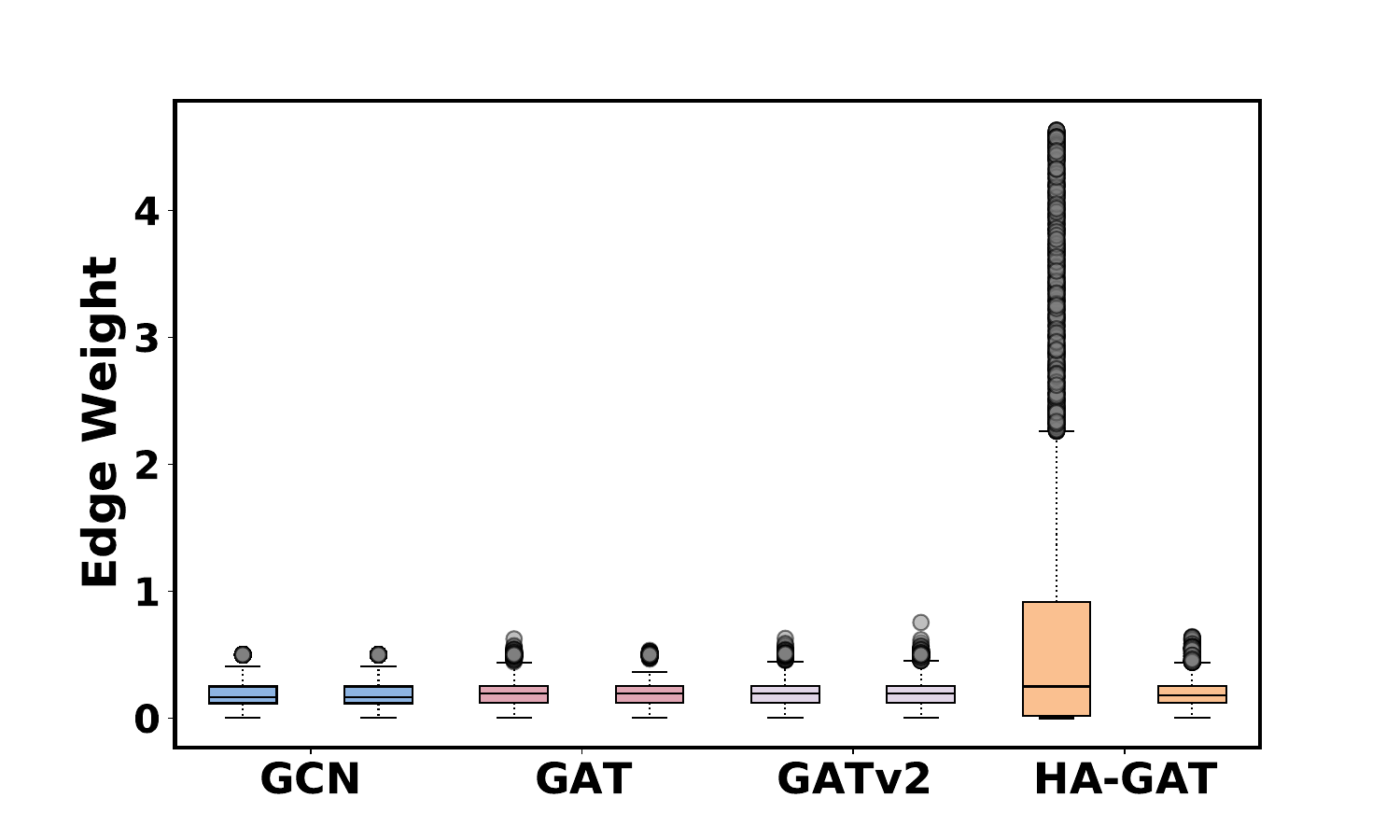}
    \caption{Results on Cora.}
    \label{fig-weight-cora}
  \end{subfigure}
  \begin{subfigure}[b]{0.45\columnwidth}
    \includegraphics[width=\linewidth]{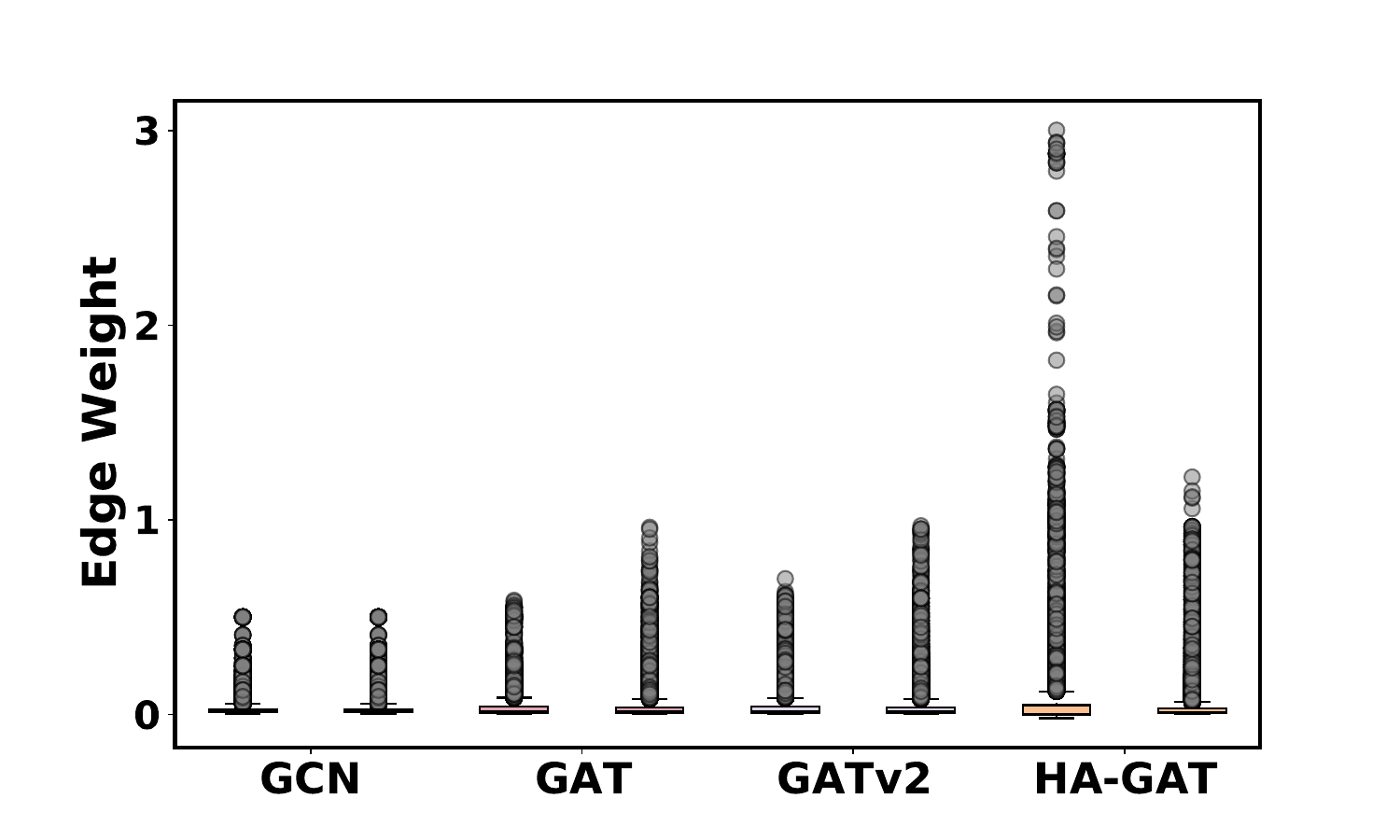}
    \caption{Results on Chameleon.}
    \label{fig-weight-chameleon}
  \end{subfigure}
  \caption{Layer-wise distributions of the edge weights learned by different models.}
  \label{fig-weight}
\end{figure}

\begin{figure}[h]
  \centering
  \begin{subfigure}[b]{0.45\columnwidth}
    \includegraphics[width=\linewidth]{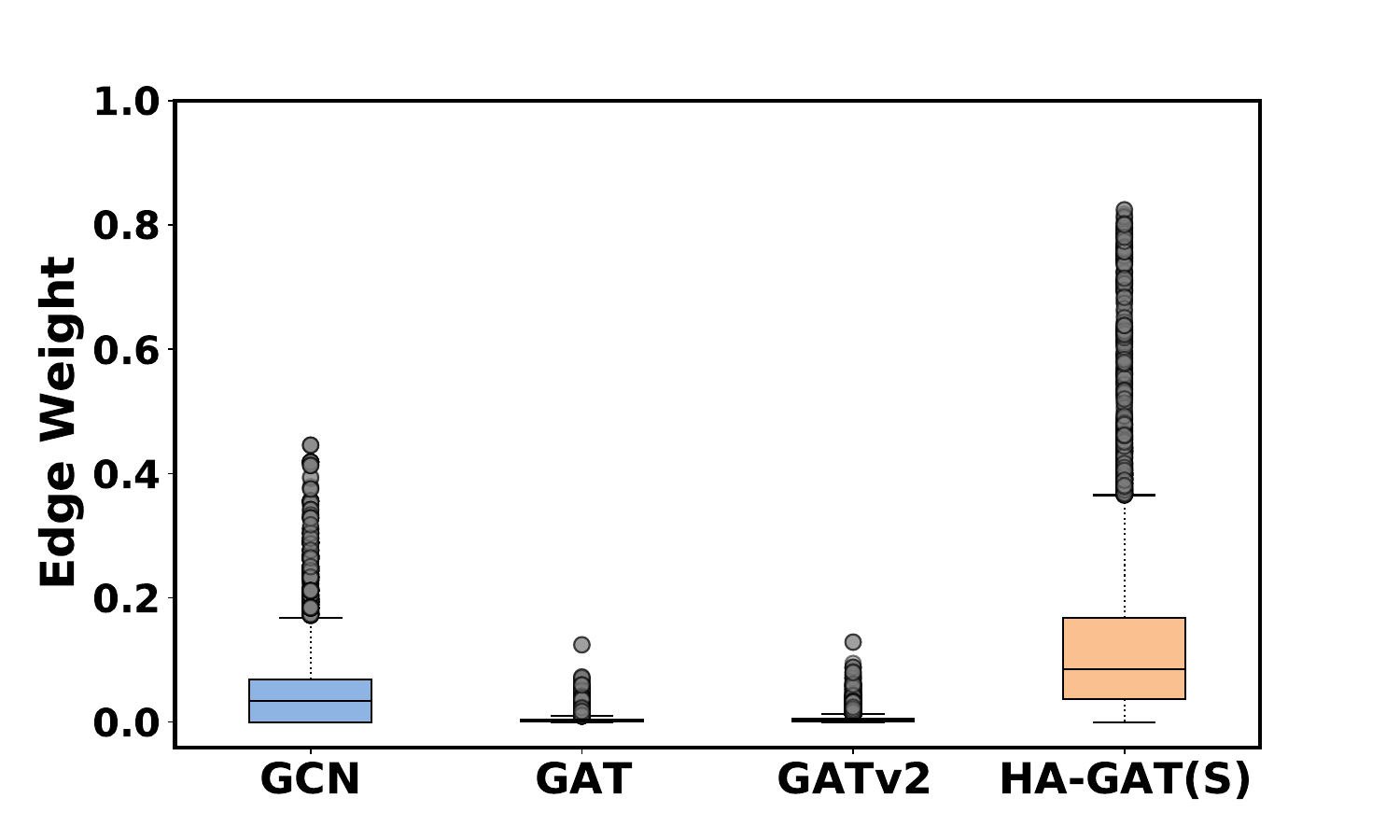}
    \caption{Results on Cora.}
    \label{fig-weight-diff-cora}
  \end{subfigure}
  \begin{subfigure}[b]{0.45\columnwidth}
    \includegraphics[width=\linewidth]{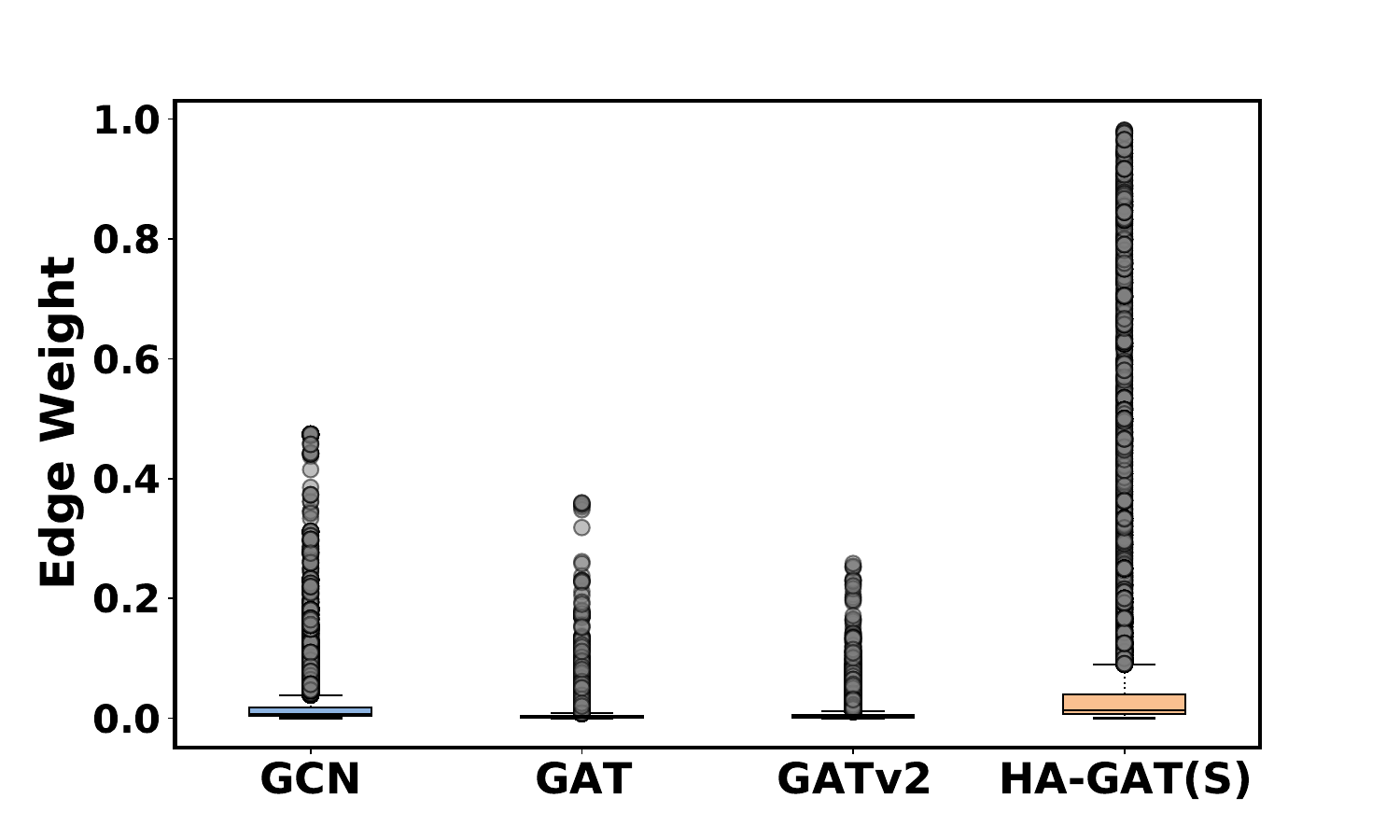}
    \caption{Results on Chameleon.}
    \label{fig-weight-diff-chameleon}
  \end{subfigure}
  \caption{Distributions of the difference between the weights learned by different models with the averaged normalization.}
  \label{fig-weight-diff}
\end{figure}

To provide an intuitive understanding, this subsection visualizes the distributions of the learned weights in our HA-GAT, two attention-based GNNs (i.e., GAT and GATv2) and a structure-based GNN (i.e., GCN).
All of these GNNs are constructed with two layers.
\cref{fig-weight} shows their layer-wise distributions of the learned weights.
As can be observed, the distributions of the learned weights in GCN, GAT, and GATv2 are similar.
Besides, there are little variations of their distributions across different layers.
On the contrary, the distributions of our HA-GAT are diverse for different layers and datasets.
Due to the employed Neighbor Norm, our HA-GAT can assign diverse weights, even more than one, which makes our attention scheme more flexible and can handle graphs with complex local distribution.

For further comparisons, we replace our normalization approach with the Softmax Norm (as shown in \cref{eq-softmax-norm}), which is employed in GAT and GATv2.
\cref{fig-weight-diff} shows the difference between the weights learned by different models with the Mean Norm (as shown in \cref{eq-mean-norm}).
GCN, which employs GCN Norm (as shown in \cref{eq-gcn-norm}) to normalize edge weight, is utilized as a baseline.
As can be observed, the weights learned by GAT and GATv2 are very close to the averaged normalization, which indicates that they assign most edges equal importance.
On the contrary, our heterophily-aware attention scheme in our HA-GAT(S) can generate diverse weights to differentiate various types of edges, which demonstrates the superiority of our attention scheme.

\begin{figure}[t]
  \centering
  \begin{subfigure}[b]{0.4\linewidth}
    \includegraphics[trim=0cm 9cm 0cm 8cm, clip, width=\linewidth]{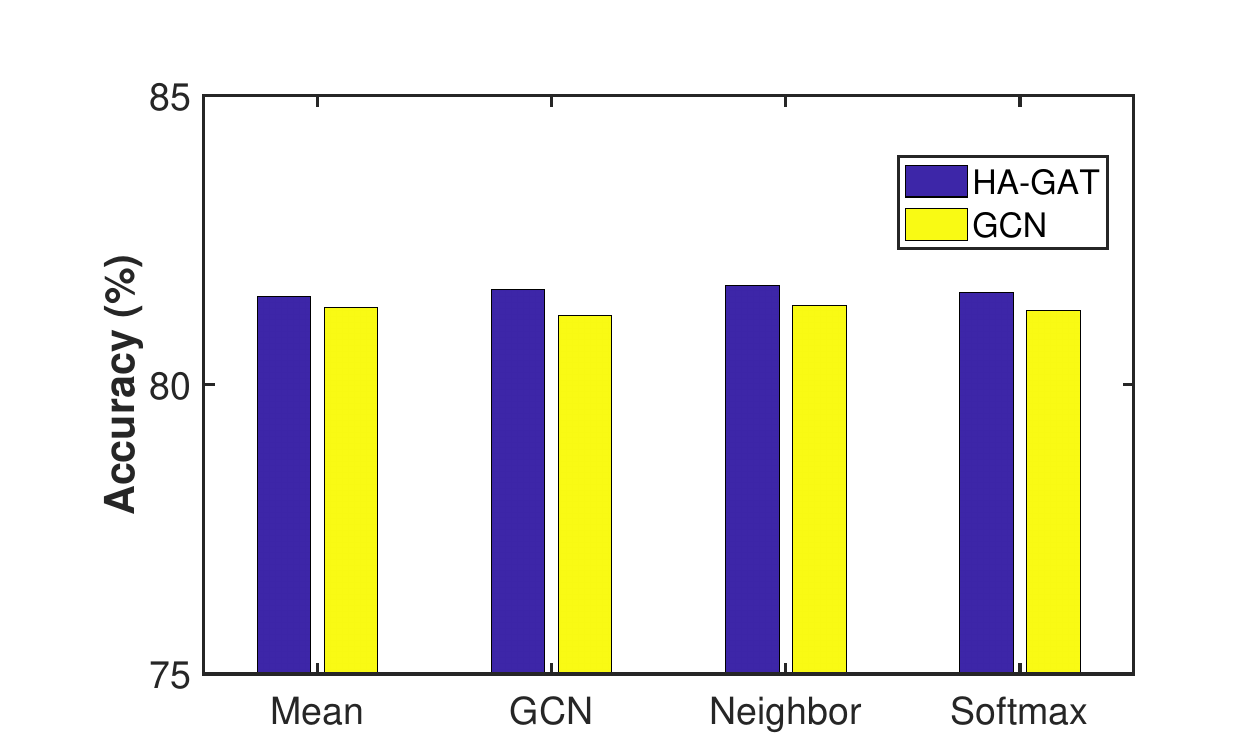}
    \caption{Results on CiteSeer.}
    \label{fig-norm-chameleon}
  \end{subfigure}
  \begin{subfigure}[b]{0.4\linewidth}
    \includegraphics[trim=0cm 9cm 0cm 8cm, clip,width=\linewidth]{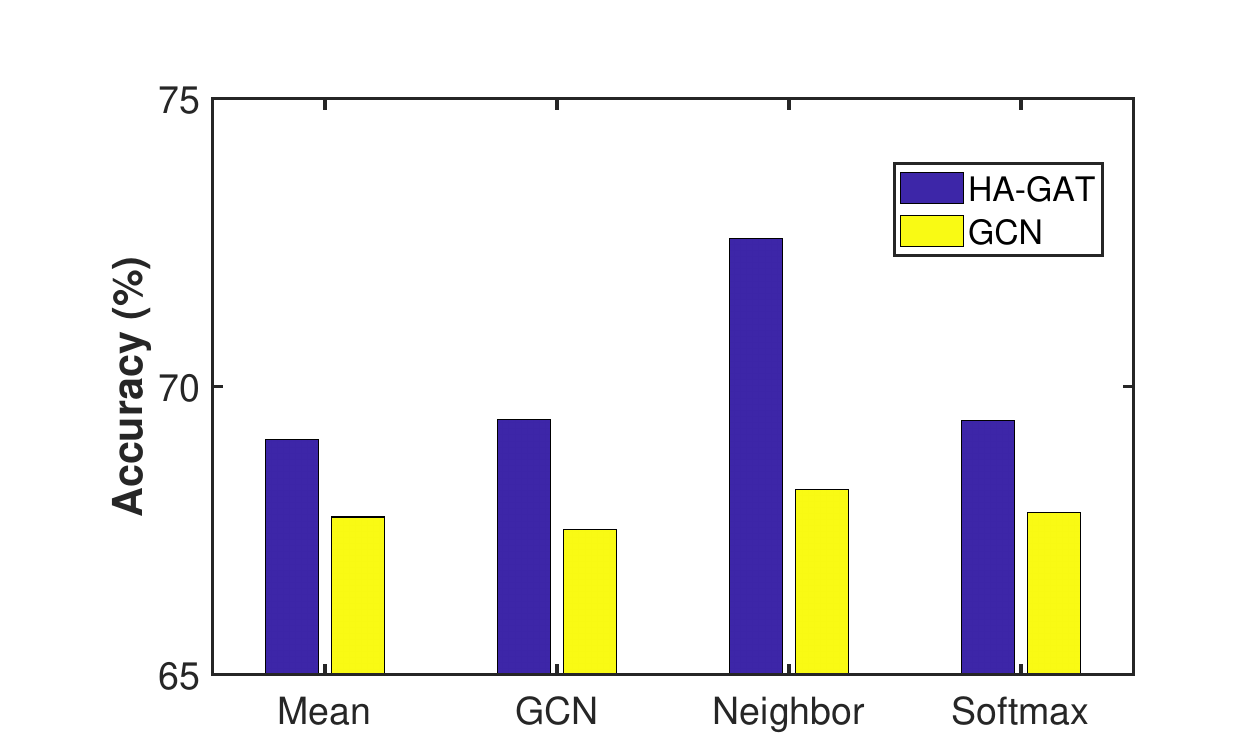}
    \caption{Results on Chameleon.}
    \label{fig-norm-cora}
  \end{subfigure}
  \caption{Results of HA-GAT and GCN with different normalization methods.}
  \label{fig-norm}
\end{figure}

\subsection{Ablation Studies of Critical Designs}
Here, we verify the effectiveness of our Neighbor Norm and the gradient scaling factor.

\subsubsection{Neighbor Norm}
\label{subsection-norm}
Our HA-GAT employs the Neighbor Norm to normalize the attention coefficients, which exploits the weighted degree of the neighbor nodes as the normalization item.
We leverage three commonly utilized normalization methods for comparison, i.e., Mean Norm, GCN Norm, and Softmax Norm.
Their details are as follows.

\noindent\textbf{Mean Norm.}
It utilizes the weighted degree of the central node $v_i$ as a normalization item, i.e., 
\begin{equation}
  \alpha_{ij}^{(l)} = \frac{w_{ij}^{(l)}}{\sum_{v_k\in\tilde{\mathcal{N}}_i} w_{ik}^{(l)}},
  \label{eq-mean-norm}
\end{equation}
where $\tilde{\mathcal{N}}_i=\mathcal{N}_{i} \cup {i}$ and each $w_{ik}^{(l)}$ is the heterophily-aware attention coefficient for edge $e_{ik}$ in the $l$-th layer, which is calculated via \cref{eq-w-1}.

\noindent\textbf{GCN Norm.}
It is utilized in GCN \cite{gcn}, which considers the weighted degrees of both the central node $v_i$ and the neighbor node $v_j$, i.e., 
\begin{equation}
  \alpha_{ij}^{(l)} = \frac{w_{ij}^{(l)}}{\sqrt{\sum_{v_k\in\tilde{\mathcal{N}}_i} w_{ik}^{(l)}}\cdot \sqrt{\sum_{v_k\in\tilde{\mathcal{N}}_j} w_{jk}^{(l)}}}.
  \label{eq-gcn-norm}
\end{equation}
Although this method is usually employed for handling undirected graphs, we exploit it here to consider the effects of both the central node and the corresponding neighbor.

\noindent\textbf{Softmax Norm.}
It utilizes the Softmax function to normalize the weights among all the edges coming into the central node and is usually used in the attention-based GNNs \cite{gat,gatv2}, i.e.,
\begin{equation}
  \alpha_{ij}^{(l)} = {\rm softmax}\left(w_{ij}\right) = \frac{{\rm exp}(w_{ij})}{\sum_{v_k\in\tilde{\mathcal{N}_i}}{\rm exp}(w_{ik})}.
  \label{eq-softmax-norm}
\end{equation}
Note that when utilizing the Softmax Norm for our HA-GAT, we eliminate the $\left(\cdot \right)_{+}$ function in \cref{eq-phi}.

As can be observed in \cref{fig-norm}, for both HA-GAT and GCN, four normalization methods achieve outstanding performances on CiteSeer, due to its simple local distributions.
However, their performances are quite different on the heterophilic graphs.
GCN and HA-GAT with Neighbor Norm outperform their corresponding variants with other normalization methods on Chameleon, which suggests that Neighbor Norm is more compatible under such circumstances.
It leads to a flexible scoring mechanism and allows diverse normalized weights (shown in \cref{appendix-weight}), thus it can be applied to more complex local distributions. 
Besides, Neighbor Norm functions decently in our heterophily-aware attention scheme.
\textit{As shown in \cref{fig-LDE}, it makes the characteristics of nodes from type T1 significant (in the first layer) and then uses them to influence other types of nodes (in the second layer).}
This observation further reveals the mechanism of our HA-GAT.

\subsubsection{Gradient Scaling Factor}
\label{appendix-lambda}
\begin{figure}[t]
  \centering
  \begin{subfigure}{.235\columnwidth}
    \includegraphics[width=\linewidth]{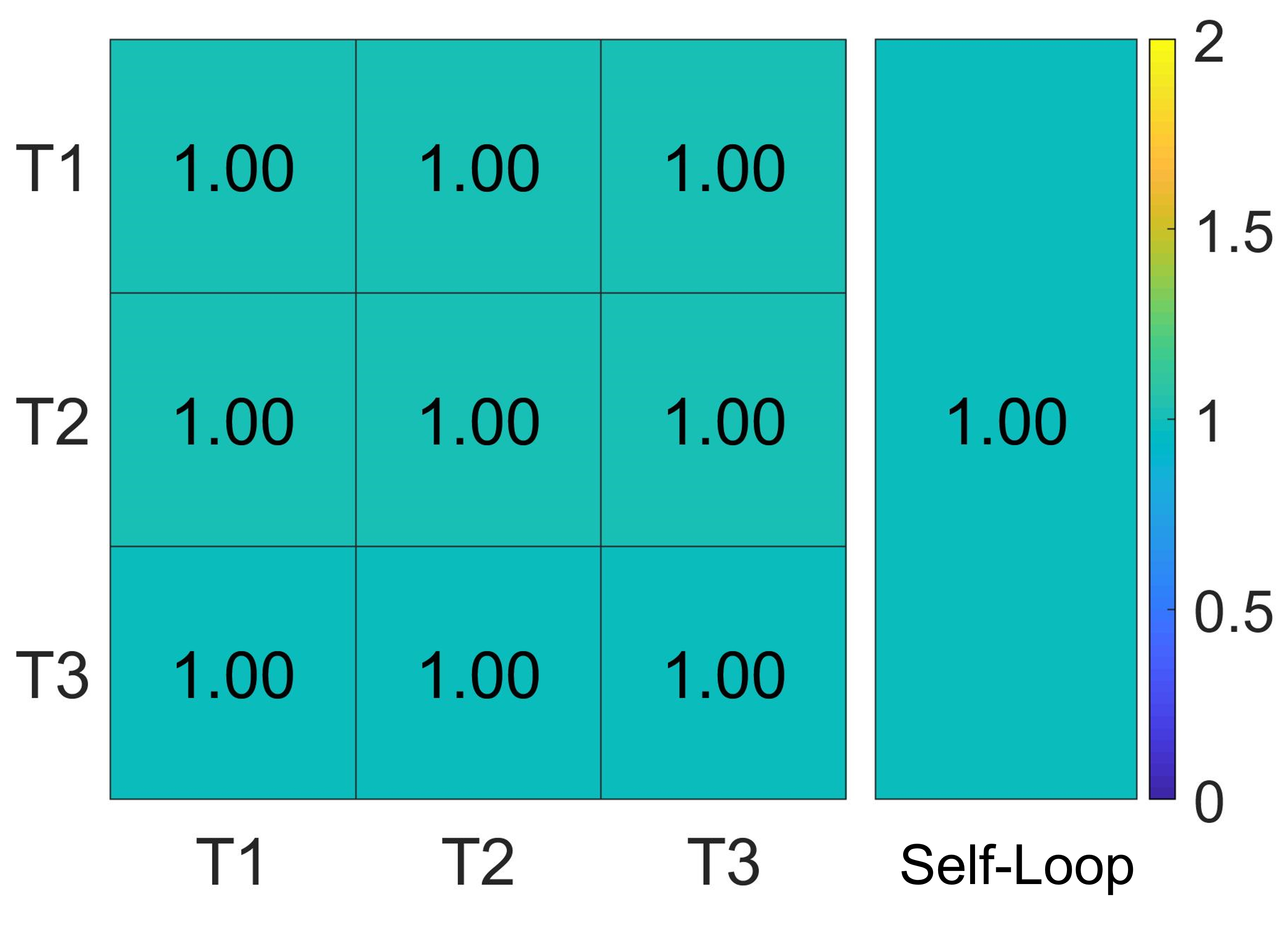}
    \caption{$\lambda$=0.00001.}
    \label{fig-r000001}
  \end{subfigure}
  \begin{subfigure}{.235\columnwidth}
    \includegraphics[width=\linewidth]{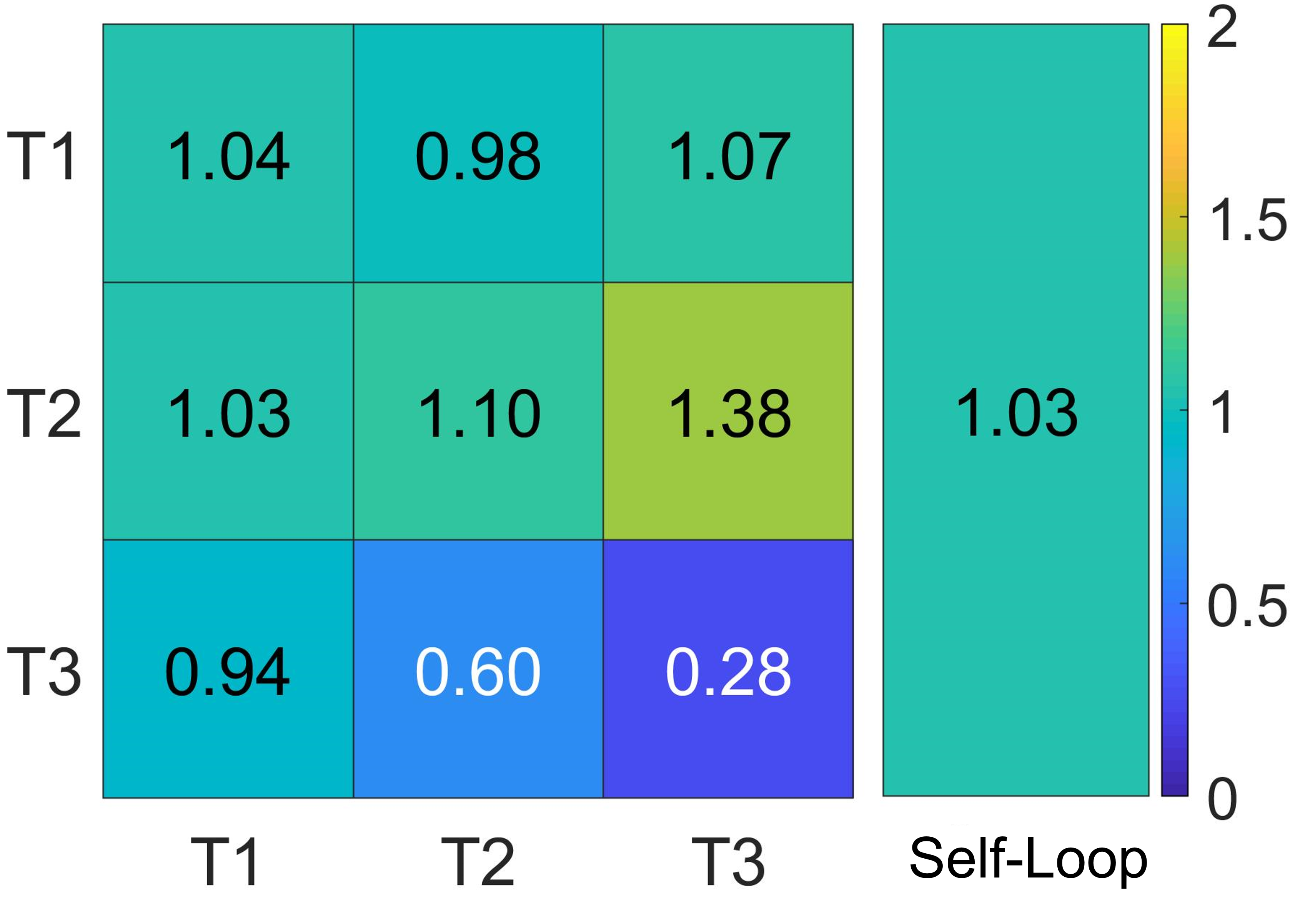}
    \caption{$\lambda$=0.1.}
    \label{fig-r01}
  \end{subfigure}
  \begin{subfigure}{.23\columnwidth}
    \includegraphics[width=\linewidth]{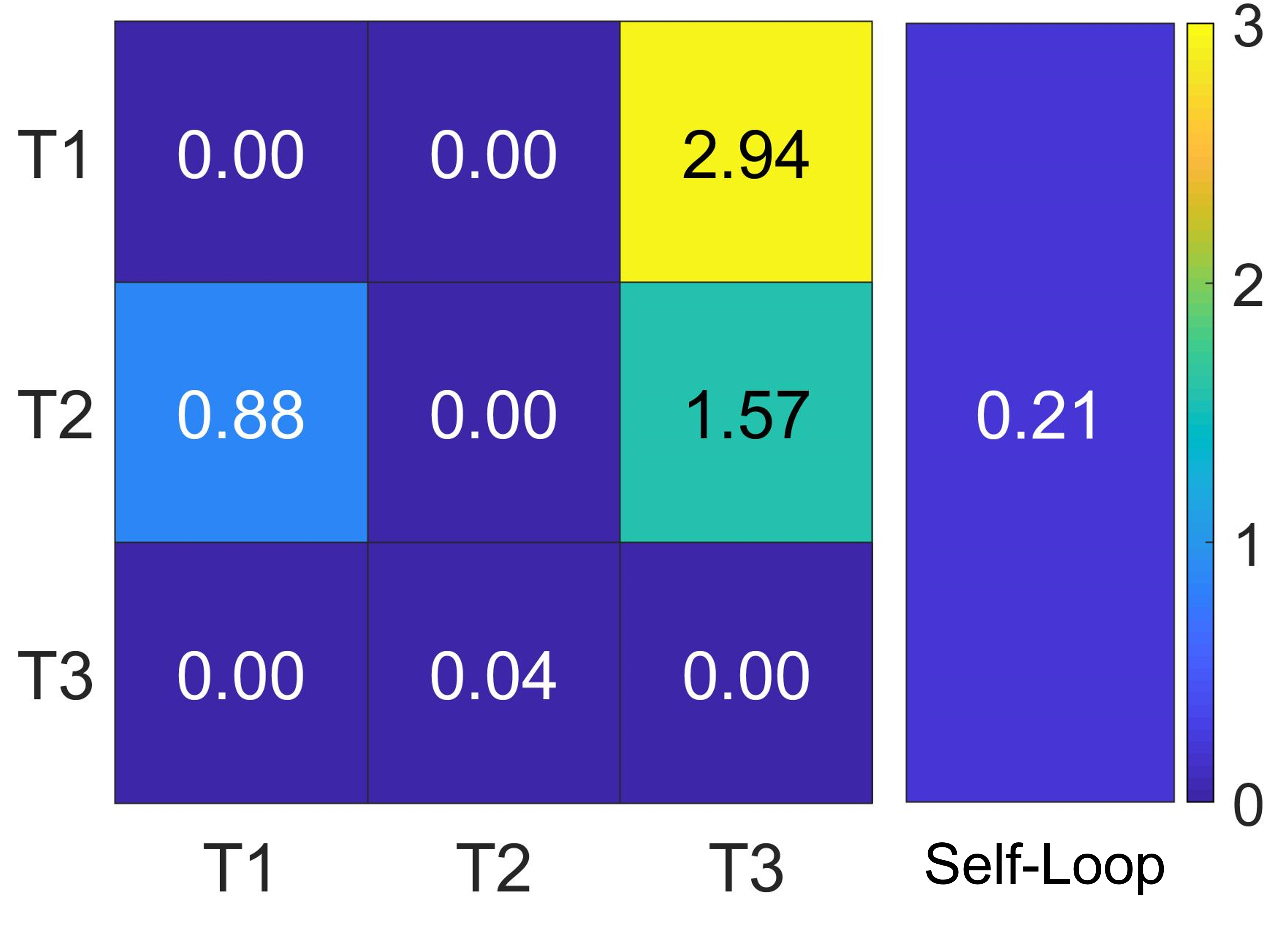}
    \caption{$\lambda$=1.}
    \label{fig-r1}
  \end{subfigure}
  \begin{subfigure}{.23\columnwidth}
    \includegraphics[width=\linewidth]{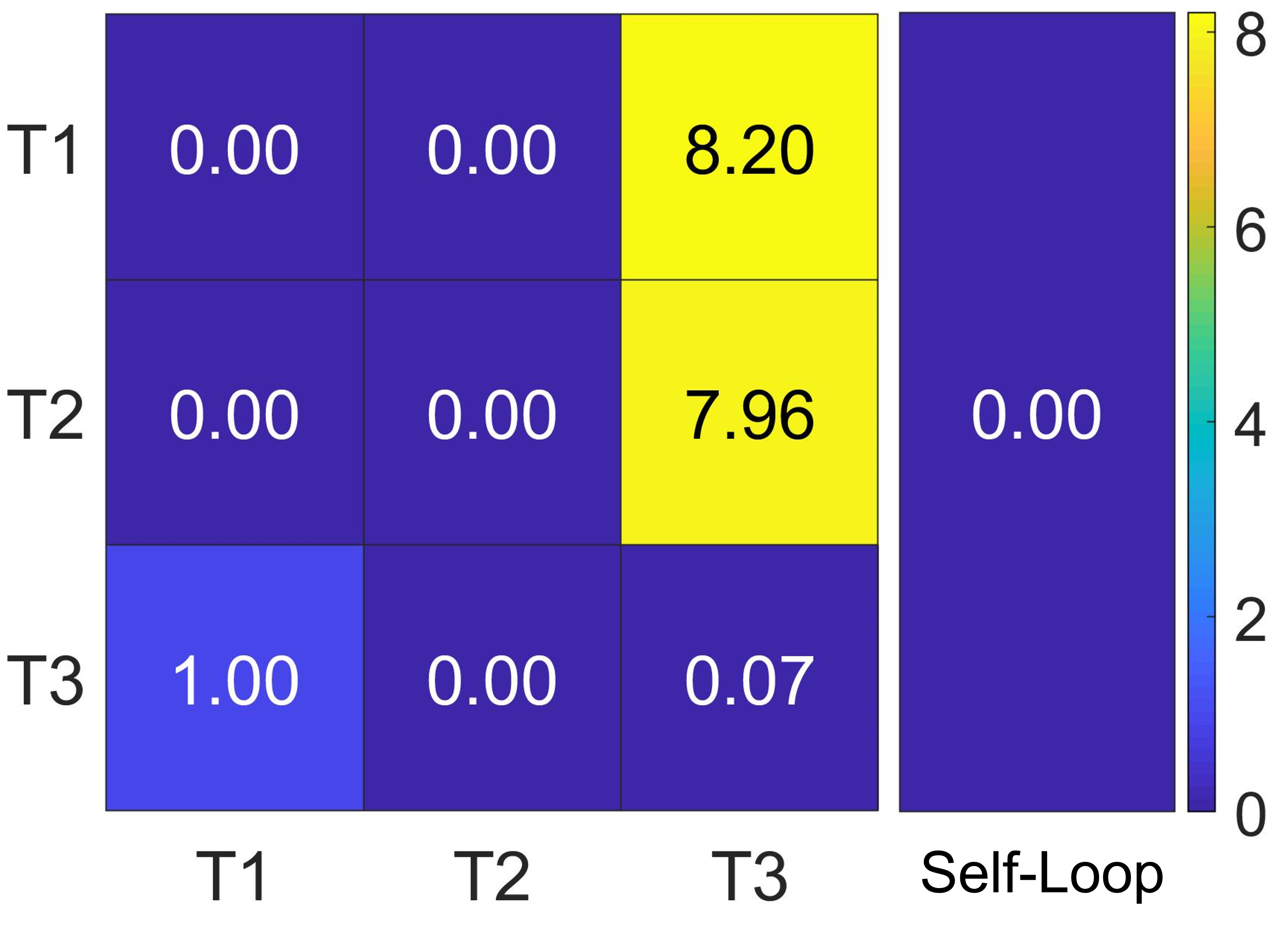}
    \caption{$\lambda$=10.}
    \label{fig-r10}
  \end{subfigure}
  \caption{LAPs of the first layer in HA-GAT with different choices of gradient scaling factor on Squirrel.}
  \label{fig-r}
\end{figure}

Here, we experimentally study the impacts of the employed gradient scaling factor, $\lambda$.
\cref{fig-r} visualizes the learned LAPs of the first layer in a 2-layered HA-GAT with different choices of gradient scaling factor on Squirrel.
Note that all the elements in LAPs are initially set as ones.
As observed, when lambda is a small value, i.e., 1e-5, the learned LAP is approximately equivalent to the initial LAP.
Then, our HA-GAT degenerates to the GCN with Neighbor Norm.
With the lambda increasing, the elements of the learned LAP are more diverse, i.e., our HA-GAT is more courageous to assign the weights of edges.
It demonstrates that the gradient scaling factor can enlarge/shrink the gradients of the parsing matrix, hence enabling our LAP to be applicable to distinct graphs.

\subsection{Parameter Study}
\begin{figure}[t]
  \centering
  \begin{subfigure}{.48\columnwidth}
    \includegraphics[width=\linewidth]{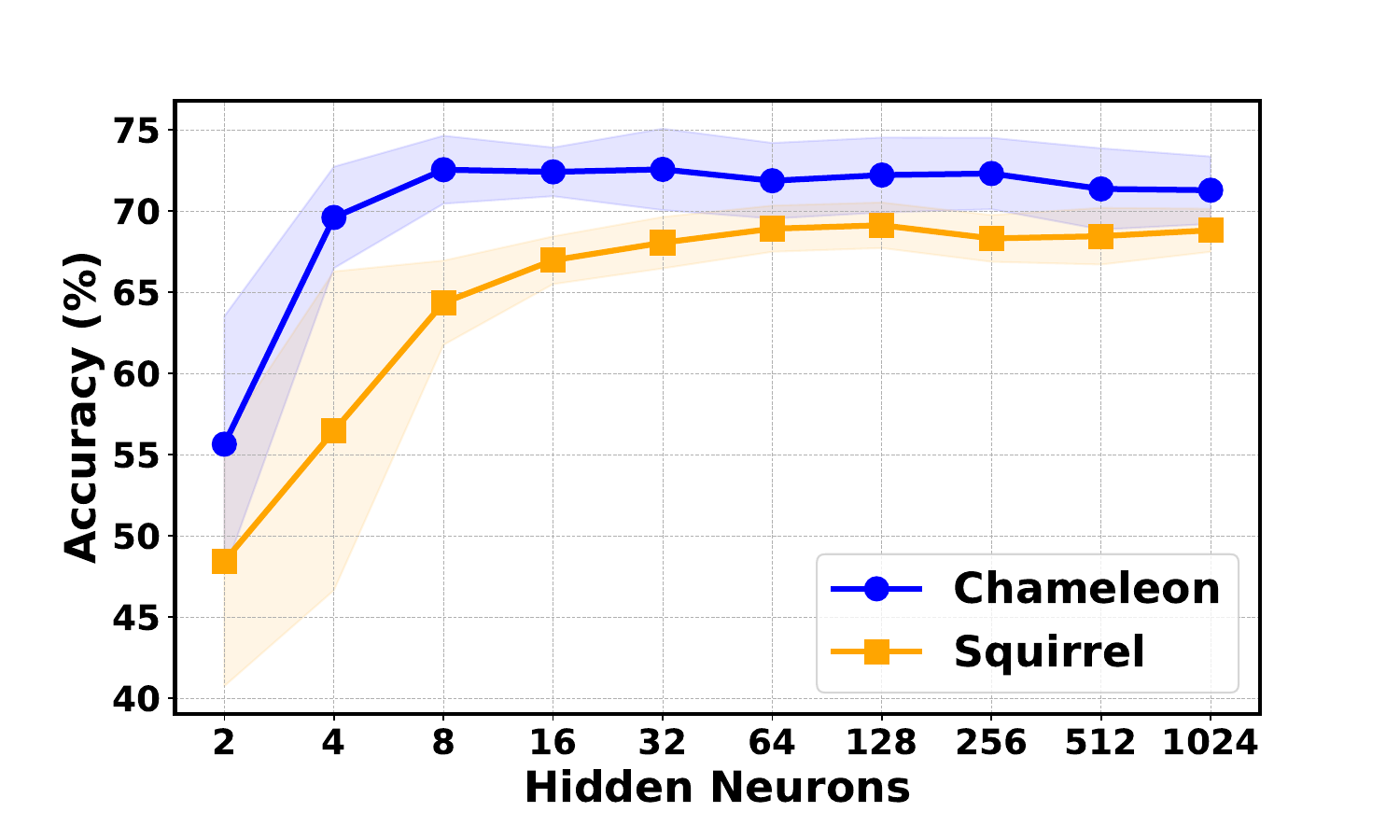}
    \caption{Performance with different Hidden Neurons.}
    \label{fig:hiddens}
  \end{subfigure}
  \begin{subfigure}{.48\columnwidth}
    \includegraphics[width=\linewidth]{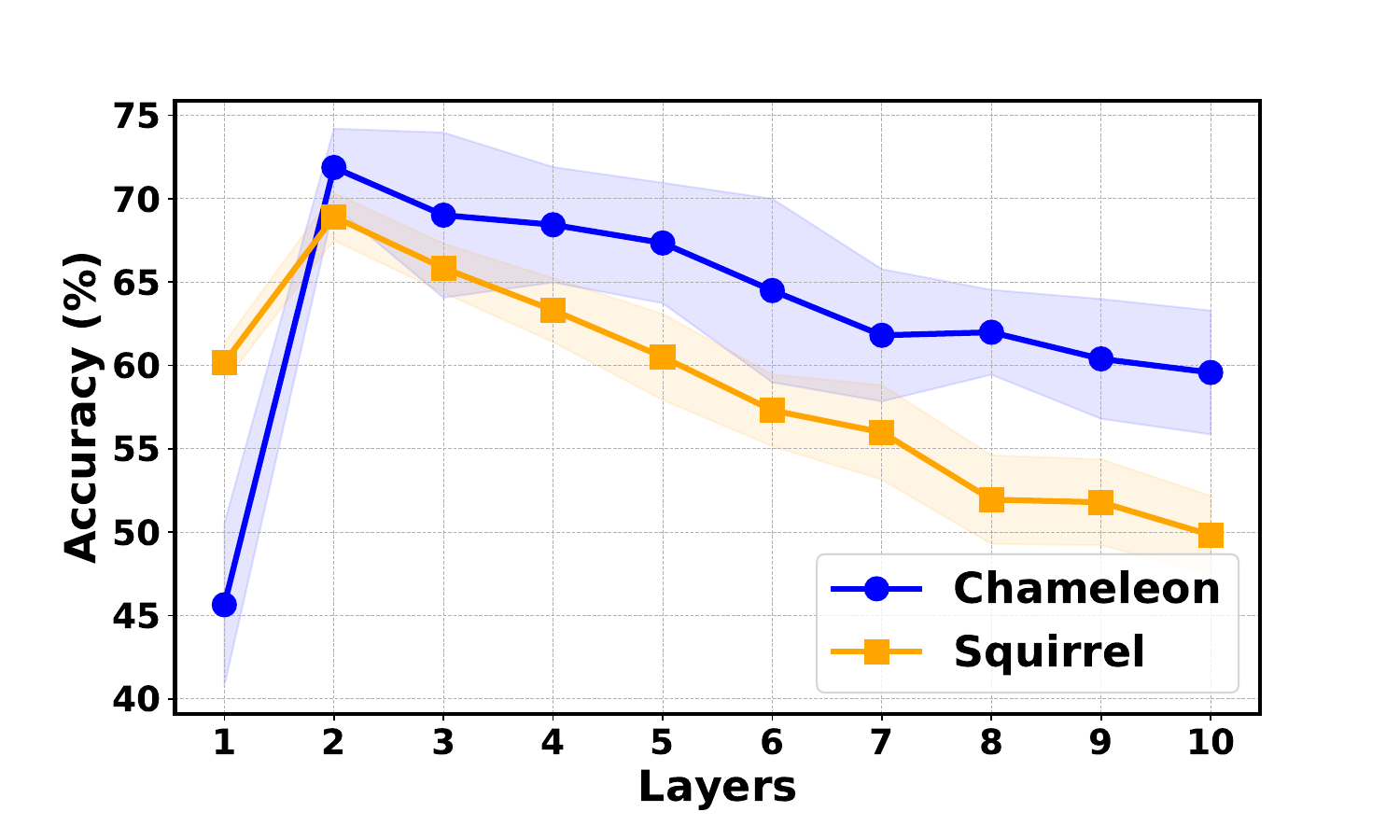}
    \caption{Performance with different Layers.}
    \label{fig:layers}
  \end{subfigure}
  \caption{Performance with different Hidden Neurons and Layers on Chameleon and Squirrel.}
  \label{fig:parameter_study}
\end{figure}

Here, we explore the impact of two critical hyperparameters: the number of hidden neurons and the number of layers within our HA-GAT.

As depicted in \cref{fig:hiddens}, our HA-GAT exhibits increased performance up to a certain number of hidden neurons. Then, its performance stabilizes at high accuracies beyond a certain threshold. This suggests a critical minimum number of hidden neurons is required for our HA-GAT to effectively learn node representations and accomplish downstream tasks. According to this observation, we adopt HA-GAT with 64 hidden neurons in our experiments.

\cref{fig:layers} displays the performance of our HA-GAT, configured with 64 hidden neurons across varying numbers of layers. It is evident that the performance of our HA-GAT significantly benefits from increasing the number of layers from 1 to 2. However, performance diminishes when the number of layers exceeds 2, leading to the over-smoothing issue \cite{deeperinsights}. This trend is consistent with findings in the broader GNN literature \cite{gcn,gat}. According to the above analysis, a 2-layered HA-GAT is employed in all our settings.

\subsection{Running Time}
\begin{table}[h]
  \scriptsize
  \centering
  \begin{tabular}{cccccc}
    \toprule
    Methods          & Actor         & Chameleon  & Squirrel  & Cornell    \\
    \midrule
    MLP              & 1.86 ± 0.13   & 1.82 ± 0.27  & 1.10 ± 0.10 & 1.03 ± 0.05  \\
    GCN              & 1.42 ± 0.02   & 6.73 ± 0.19  & 1.93 ± 0.04 & 2.17 ± 0.85  \\
    GAT              & 2.52 ± 0.32   & 3.08 ± 0.55  & 3.72 ± 0.62 & 2.85 ± 0.45  \\
    GRP-GNN          & 4.13 ± 0.17   & 3.67 ± 0.07  & 4.80 ± 1.19 & 3.89 ± 0.08  \\
    DMP              & 20.52 ± 0.49  & M            & M         & 14.65 ± 1.32 \\
    BM-GCN           & 115.13 ± 1.28 & 26.16 ± 5.85 & M         & 21.01 ± 4.09 \\
    HA-GAT           & 5.45 ± 0.75   & 4.20 ± 0.69  & 9.94 ± 3.83 & 3.72 ± 0.17  \\
    \bottomrule
  \end{tabular}
  \caption{Running times (s) of the supervised node classification tasks. (Note that M denotes that it occurs an out-of-memory error on our Nvidia GeForce RTX 2080Ti GPU.)}
  \label{table-run-time}
\end{table}

\cref{table-run-time} presents the running times (s) of experiments in \cref{table-supervised}.
All the experiments are conducted on our Nvidia GeForce RTX 2080Ti GPU.
In our settings, all the following methods except BM-GCN are trained for a maximum of 1000 epochs with an early stopping condition at 200 epochs.
For BM-GCN, we follow their 2-staged learning strategy, which contains 400 and 1600 epochs for its two training stages.
As can be observed, the reported running times of our HA-GAT match the complexity analysis.

\section{Conclusion}
This paper firstly reveals the benefits of modeling the edge heterophily for handling the heterophily problem, i.e., if a GNN assigns different weights to edges according to different heterophilic types, it can learn effective local attention patterns, enabling nodes to acquire appropriate information from distinct neighbors.
Then, by fully exploring and utilizing the local distribution as the underlying heterophily, a novel Heterophily-Aware Graph Attention Network (HA-GAT) is proposed to handle the networks with different homophily ratios.
Extensive results have demonstrated that our HA-GAT achieves state-of-the-art performances on eight datasets with different homophily ratios.
To further demonstrate the effectiveness of the proposed HA-GAT, we analyze the proposed heterophily-aware attention scheme and local distribution exploration and seek for an interpretation from their mechanism.
More interpretations of the learned weights will require domain knowledge under study and are left as future work.
Consequently, we believe our work can be applied in various areas, such as social analysis \cite{wang2018exploiting,wang2017onymity}.

\section*{Acknowledgement}
This work was supported in part by the National Natural Science Foundation of China under Grants 62272020, U20B2069, and 62376088, in part by State Key Laboratory of Software Development Environment (SKLSDE-2023ZX-16), in part by Hebei Natural Science Foundation No. F2024202047, in part by the Outstanding Research Project of Shen Yuan Honors College, BUAA under Grant 230123201, and in part by the Fundamental Research Funds for the Central Universities.

\bibliographystyle{elsarticle-num} 
\bibliography{hagat_pr.bib}

\begin{thebibliography}{10}
\expandafter\ifx\csname url\endcsname\relax
  \def\url#1{\texttt{#1}}\fi
\expandafter\ifx\csname urlprefix\endcsname\relax\def\urlprefix{URL }\fi
\expandafter\ifx\csname href\endcsname\relax
  \def\href#1#2{#2} \def\path#1{#1}\fi

\bibitem{gcn}
T.~N. Kipf, M.~Welling, Semi-supervised classification with graph convolutional
  networks, in: {ICLR}, 2017.

\bibitem{gat}
P.~Velickovic, G.~Cucurull, A.~Casanova, A.~Romero, P.~Li{\`{o}}, Y.~Bengio,
  Graph attention networks, in: {ICLR}, 2018.

\bibitem{graphsage}
W.~L. Hamilton, Z.~Ying, J.~Leskovec, Inductive representation learning on
  large graphs, in: {NIPS}, 2017, pp. 1024--1034.

\bibitem{bi-gcn}
J.~Wang, Y.~Wang, Z.~Yang, L.~Yang, Y.~Guo, Bi-gcn: Binary graph convolutional
  network, in: {IEEE} {CVPR}, 2021, pp. 1561--1570.

\bibitem{bai2021learning}
L.~Bai, Y.~Jiao, L.~Cui, L.~Rossi, Y.~Wang, S.~Y. Philip, E.~R. Hancock,
  Learning graph convolutional networks based on quantum vertex information
  propagation, IEEE Transactions on Knowledge and Data Engineering 35~(2)
  (2021) 1747--1760.

\bibitem{cui2021learning}
L.~Cui, L.~Bai, X.~Bai, Y.~Wang, E.~R. Hancock, Learning aligned vertex
  convolutional networks for graph classification, IEEE Transactions on Neural
  Networks and Learning Systems (2021).

\bibitem{regnn}
J.~Wang, Y.~Guo, L.~Yang, Y.~Wang, Enabling homogeneous gnns to handle
  heterogeneous graphs via relation embedding, IEEE Transactions on Big Data
  9~(6) (2023) 1697--1710.

\bibitem{bigcn2}
J.~Wang, Y.~Guo, L.~Yang, Y.~Wang, Binary graph convolutional network with
  capacity exploration, IEEE Transactions on Pattern Analysis and Machine
  Intelligence 46~(5) (2024) 3031--3046.

\bibitem{gnn_app_social1}
J.~Qiu, J.~Tang, H.~Ma, Y.~Dong, K.~Wang, J.~Tang, Deepinf: Social influence
  prediction with deep learning, in: {ACM} {SIGKDD}, 2018, pp. 2110--2119.

\bibitem{gnn_app_object_tracking1}
X.~Weng, Y.~Wang, Y.~Man, K.~M. Kitani, {GNN3DMOT:} graph neural network for 3d
  multi-object tracking with 2d-3d multi-feature learning, in: {IEEE} {CVPR},
  2020, pp. 6498--6507.

\bibitem{nlp2}
D.~Marcheggiani, I.~Titov, Encoding sentences with graph convolutional networks
  for semantic role labeling, in: {EMNLP}, 2017, pp. 1506--1515.

\bibitem{message-passing}
J.~Gilmer, S.~S. Schoenholz, P.~F. Riley, O.~Vinyals, G.~E. Dahl, Neural
  message passing for quantum chemistry, in: {ICML}, 2017, pp. 1263--1272.

\bibitem{gtn}
S.~Yun, M.~Jeong, R.~Kim, J.~Kang, H.~J. Kim, Graph transformer networks, in:
  {NIPS}, 2019, pp. 11983--11993.

\bibitem{wijesinghe2021new}
A.~Wijesinghe, Q.~Wang, A new perspective on" how graph neural networks go
  beyond weisfeiler-lehman?", in: {ICLR}, 2022.

\bibitem{GAT-cosine}
K.~K. Thekumparampil, C.~Wang, S.~Oh, L.-J. Li, Attention-based graph neural
  network for semi-supervised learning, arXiv preprint arXiv:1803.03735 (2018).

\bibitem{Gaan}
J.~Zhang, X.~Shi, J.~Xie, H.~Ma, I.~King, D.-Y. Yeung, Gaan: Gated attention
  networks for learning on large and spatiotemporal graphs, arXiv preprint
  arXiv:1803.07294 (2018).

\bibitem{SuperGAT}
D.~Kim, A.~Oh, How to find your friendly neighborhood: Graph attention design
  with self-supervision, in: {ICLR}, 2021.

\bibitem{CS-GNN}
Y.~Hou, J.~Zhang, J.~Cheng, K.~Ma, R.~T. Ma, H.~Chen, M.-C. Yang, Measuring and
  improving the use of graph information in graph neural networks, in: {ICLR},
  2020.

\bibitem{gatv2}
S.~Brody, U.~Alon, E.~Yahav, How attentive are graph attention networks?, in:
  {ICLR}, 2022.

\bibitem{H2GCN}
J.~Zhu, Y.~Yan, L.~Zhao, M.~Heimann, L.~Akoglu, D.~Koutra, Beyond homophily in
  graph neural networks: Current limitations and effective designs, in: {NIPS},
  Vol.~33, 2020, pp. 7793--7804.

\bibitem{geom-gcn}
H.~Pei, B.~Wei, K.~C.-C. Chang, Y.~Lei, B.~Yang, Geom-gcn: Geometric graph
  convolutional networks, in: {ICLR}, 2020.

\bibitem{DMP}
L.~Yang, M.~Li, L.~Liu, C.~Wang, X.~Cao, Y.~Guo, Diverse message passing for
  attribute with heterophily, in: {NIPS}, Vol.~34, 2021, pp. 4751--4763.

\bibitem{FAGCN}
D.~Bo, X.~Wang, C.~Shi, H.~Shen, Beyond low-frequency information in graph
  convolutional networks, in: {AAAI}, Vol.~35, 2021, pp. 3950--3957.

\bibitem{BM-GCN}
D.~He, C.~Liang, H.~Liu, M.~Wen, P.~Jiao, Z.~Feng, Block modeling-guided graph
  convolutional neural networks, in: {AAAI}, Vol.~36, 2022, pp. 4022--4029.

\bibitem{cpgnn}
J.~Zhu, R.~A. Rossi, A.~Rao, T.~Mai, N.~Lipka, N.~K. Ahmed, D.~Koutra, Graph
  neural networks with heterophily, in: {AAAI}, Vol.~35, 2021, pp.
  11168--11176.

\bibitem{ma2021homophily}
Y.~Ma, X.~Liu, N.~Shah, J.~Tang, Is homophily a necessity for graph neural
  networks?, in: {ICLR}, 2022.

\bibitem{wang2024understanding}
J.~Wang, Y.~Guo, L.~Yang, Y.~Wang, Understanding heterophily for graph neural
  networks, arXiv preprint arXiv:2401.09125 (2024).

\bibitem{MWGNN}
X.~Ma, Q.~Chen, Y.~Ren, G.~Song, L.~Wang, Meta-weight graph neural network:
  Push the limits beyond global homophily, in: {WWW}, 2022, pp. 1270--1280.

\bibitem{acm-gnn}
S.~Luan, C.~Hua, Q.~Lu, J.~Zhu, M.~Zhao, S.~Zhang, X.-W. Chang, D.~Precup,
  Revisiting heterophily for graph neural networks, in: {NeurIPS}, Vol.~35,
  2022, pp. 1362--1375.

\bibitem{gbkgnn}
L.~Du, X.~Shi, Q.~Fu, X.~Ma, H.~Liu, S.~Han, D.~Zhang, Gbk-gnn: Gated bi-kernel
  graph neural networks for modeling both homophily and heterophily, in: {WWW},
  2022, pp. 1550--1558.

\bibitem{WRGNN}
S.~Suresh, V.~Budde, J.~Neville, P.~Li, J.~Ma, Breaking the limit of graph
  neural networks by improving the assortativity of graphs with local mixing
  patterns, in: {ACM} {SIGKDD}, 2021, pp. 1541--1551.

\bibitem{gcn2}
M.~Defferrard, X.~Bresson, P.~Vandergheynst, Convolutional neural networks on
  graphs with fast localized spectral filtering, in: {NIPS}, 2016, pp.
  253--261.

\bibitem{mixhop}
S.~Abu{-}El{-}Haija, B.~Perozzi, A.~Kapoor, N.~Alipourfard, K.~Lerman,
  H.~Harutyunyan, G.~V. Steeg, A.~Galstyan, Mixhop: Higher-order graph
  convolutional architectures via sparsified neighborhood mixing, in: {ICML},
  2019, pp. 21--29.

\bibitem{gpr-gnn}
E.~Chien, J.~Peng, P.~Li, O.~Milenkovic, Adaptive universal generalized
  pagerank graph neural network, in: {ICLR}, 2021.

\bibitem{non-local-gnn}
M.~Liu, Z.~Wang, S.~Ji, Non-local graph neural networks, IEEE Transactions on
  Pattern Analysis and Machine Intelligence (2021) 1--1.

\bibitem{GPNN}
T.~Yang, Y.~Wang, Z.~Yue, Y.~Yang, Y.~Tong, J.~Bai, Graph pointer neural
  networks, arXiv preprint arXiv:2110.00973 (2021).

\bibitem{UGCN}
D.~Jin, Z.~Yu, C.~Huo, R.~Wang, X.~Wang, D.~He, J.~Han, Universal graph
  convolutional networks, in: {NIPS}, Vol.~34, 2021, pp. 10654--10664.

\bibitem{glognn}
X.~Li, R.~Zhu, Y.~Cheng, C.~Shan, S.~Luo, D.~Li, W.~Qian, Finding global
  homophily in graph neural networks when meeting heterophily, in: {ICML}, Vol.
  162, 2022, pp. 13242--13256.

\bibitem{citation}
P.~Sen, G.~Namata, M.~Bilgic, L.~Getoor, B.~Galligher, T.~Eliassi-Rad,
  Collective classification in network data, AI magazine 29~(3) (2008) 93--93.

\bibitem{planetoid}
Z.~Yang, W.~W. Cohen, R.~Salakhutdinov, Revisiting semi-supervised learning
  with graph embeddings, in: {ICML}, 2016, pp. 40--48.

\bibitem{jk-net}
K.~Xu, C.~Li, Y.~Tian, T.~Sonobe, K.-i. Kawarabayashi, S.~Jegelka,
  Representation learning on graphs with jumping knowledge networks, in:
  {ICML}, 2018, pp. 5453--5462.

\bibitem{adam}
D.~P. Kingma, J.~Ba, Adam: {A} method for stochastic optimization, in: {ICLR},
  2015.

\bibitem{deeperinsights}
Q.~Li, Z.~Han, X.~Wu, Deeper insights into graph convolutional networks for
  semi-supervised learning, in: {AAAI}, 2018, pp. 3538--3545.

\bibitem{wang2018exploiting}
Z.~Wang, M.~Jusup, L.~Shi, J.-H. Lee, Y.~Iwasa, S.~Boccaletti, Exploiting a
  cognitive bias promotes cooperation in social dilemma experiments, Nature
  communications 9~(1) (2018) 2954.

\bibitem{wang2017onymity}
Z.~Wang, M.~Jusup, R.-W. Wang, L.~Shi, Y.~Iwasa, Y.~Moreno, J.~Kurths, Onymity
  promotes cooperation in social dilemma experiments, Science advances 3~(3)
  (2017) e1601444.

\end{thebibliography}

\end{document}